%% file: main.tex
\documentclass[10pt]{article}
\usepackage[margin=1in]{geometry}
\usepackage{mathpazo}
\usepackage{booktabs}

\usepackage[utf8]{inputenc} 
\usepackage{url}            
\usepackage{amsfonts}       
\usepackage{nicefrac}       
\usepackage{microtype}      
\usepackage[dvipsnames]{xcolor}

\usepackage{bbm}
\usepackage{amsmath,amssymb,amsthm}
\usepackage{xspace}
\usepackage{graphicx}
\usepackage{color}
\usepackage[colorlinks=true,linkcolor=blue,citecolor=blue,urlcolor=blue]{hyperref}
\usepackage[textsize=tiny]{todonotes}
\usepackage{subcaption}
\usepackage{comment}
\usepackage{algorithm,algpseudocode}
\usepackage[T1]{fontenc}
\usepackage[backend=biber,style=alphabetic,natbib=true,maxbibnames=99,minalphanames=3]{biblatex}
\usepackage{makecell}
\usepackage[normalem]{ulem}

\usepackage{mathtools}

\newcommand{\edit}[1]{\textcolor{ForestGreen}{#1}}

\newcommand{\josh}[1]{\todo[color=blue!40]{JV: #1}}

\title{Dataset Interfaces: Diagnosing Model Failures Using Controllable Counterfactual Generation}
\author{
    Joshua Vendrow\thanks{Equal contribution.} \\
	MIT \\
	\texttt{jvendrow@mit.edu} \\
	\and
	Saachi Jain\footnotemark[1] \\
	MIT \\
	\texttt{saachij@mit.edu}
    \and
    Logan Engstrom \\
	MIT \\
	\texttt{engstrom@mit.edu}
    \and
	Aleksander M\k{a}dry \\
	MIT \\
    \texttt{madry@mit.edu}
}
\date{}
\makeatletter
\renewcommand{\paragraph}{%
  \@startsection{paragraph}{4}%
  {\z@}{1.5ex \@plus 1ex \@minus .2ex}{-1em}%
  {\normalfont\normalsize\bfseries}%
}
\makeatother

\begin{document}
    \maketitle
    \begin{abstract}
        \input{sections/abstract}
    \end{abstract}

    \section{Introduction}
    \label{sec:intro}
    \input{sections/intro}

    \section{Dataset Interfaces: Unifying methods for Counterfactual Generation}
    \label{sec:data_interfaces}
    \input{sections/dataset_interfaces.tex}

    \input{sections/dataset_interfaces_2.tex}

    \section{Generating dataset-specific counterfactual examples}
    \label{sec:method}
    \input{sections/method}

    \section{Evaluating the Generated Counterfactual Examples}
    \label{sec:distribution}
    \input{sections/distributions}

    \section{Fine-Grained Model Debugging}
    \label{sec:debugging}
    \input{sections/debugging}

    \section{Evaluating Distribution Shift Robustness}
    \label{sec:benchmark}
    \input{sections/benchmark}

    \section{Related Work}
    \label{sec:related_work}
    \input{sections/related_work}

    \section{Conclusion}
    \label{sec:conclusion}
    \input{sections/conclusion}

    \section{Acknowledgements}
    \input{sections/acks}
 
    \printbibliography

    \clearpage
    \appendix
    \input{sections/appendix}

\end{document}

%% file: sections/abstract.tex
Distribution shift is a major source of failure for machine learning models. However, evaluating model reliability under distribution shift can be challenging, especially since it may be difficult to acquire \textit{counterfactual examples} that exhibit a specified shift. In this work, we introduce the notion of a \textit{dataset interface}: a framework that, given an input dataset and a user-specified shift, returns instances from that input distribution that exhibit the desired shift. We study a number of natural implementations for such an interface, and find that they often introduce confounding shifts that complicate model evaluation. Motivated by this, we propose a dataset interface implementation that leverages Textual Inversion to tailor generation to the input distribution.
We then demonstrate how applying this dataset interface to the ImageNet dataset enables studying model behavior across a diverse array of distribution shifts, including variations in background, lighting, and attributes of the objects.\footnote{Code available at \url{https://github.com/MadryLab/dataset-interfaces}.}

%% file: sections/intro.tex
Suppose we would like to deploy a vision model (for example, one trained on ImageNet). Naturally, we  would like this model to perform reliably in a variety of real-world contexts and, especially, with respect to any of the (inevitable) corner cases, i.e., real-world inputs that are underrepresented in the training dataset. 
Indeed, we have ample evidence that machine learning models can fail when facing so-called distribution shift, including changes of the background \cite{beery2018recognition, xiao2020noise, barbu2019objectnet} and object pose \cite{engstrom2019rotation, alcorn2019strike}, as well as variability in data collection pipelines \cite{recht2018imagenet, engstrom2020identifying}. 

\begin{figure}[t!]
    \centering 
    \includegraphics[width=1\columnwidth]{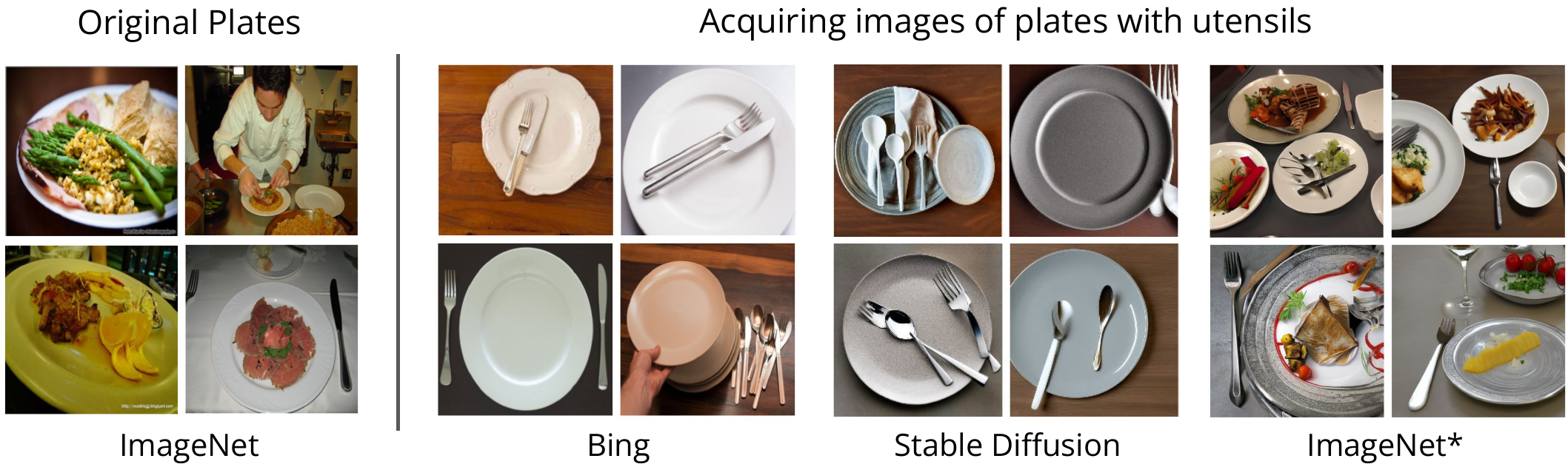}
    \caption{
    Acquiring ImageNet counterfactual examples of ``plates with utensils'' using the Bing search engine, Stable Diffusion, or our dataset interface ImageNet$*$. The Bing and Stable Diffusion plates are almost exclusively empty, and thus do not fully match the images in ImageNet (as the latter often contain food). In contrast, ImageNet$*$ can generate counterfactual examples of ``plates with utensils'' that match the ImageNet distribution much more closely. 
    }
    \vspace{-1em}
    \label{fig:bing_figure}
\end{figure}

How can we then ensure that our model will continue to perform well in new environments? To make this more concrete, suppose that our hypothetical ImageNet-trained model is being deployed to identify objects (such as dogs, chairs, plates, etc.) in images. We would like to make sure that the model will perform well regardless of the object's type (e.g., ``brown dog'', ``ceramic plate''), condition (e.g., ``dog with harness,'' ``empty plate''), and setting (e.g., ``dog on a beach,'' ``plate at a picnic'').
One powerful primitive for assessing our model's performance in such scenarios is \textit{counterfactual generation} --- acquiring images (\textit{counterfactual examples}) that conform to the training distribution except for exhibiting a specified change.
So, for instance, to test the model's performance on ``plate with utensils,'' we might want to evaluate our model on images of plates that match the distribution of ImageNet (e.g., in terms of background, camera angle, zoom, plate style) except for the fact that the plates have utensils alongside them. But how can we acquire such counterfactual examples? After all, without access to the original data-generating process, collecting new examples in a specified context can be challenging.

Currently, practitioners use a few natural methods for generating counterfactual
examples. Returning to our example of ``plate with utensils,'' one (labor
intensive) strategy is to manually take photographs of ImageNet-style scenes of
plates with and without utensils. More scalable approaches include scraping images
of plates with utensils from the internet (e.g., using Google or Bing search
engines), or synthetically generating such images (e.g., using a text-to-image
diffusion model such as Stable Diffusion~\cite{rombach2022high}). However, images produced using such approaches often reflect additional \textit{confounding shifts}. For example, querying the Bing
search engine or Stable Diffusion with the prompt ``a photo of a plate with utensils''
surfaces plates that are almost exclusively empty, while in ImageNet the plates
usually contain food (see Figure~\ref{fig:bing_figure}). Any observed change in
the model's accuracy on these images might thus well be due to this confounding
shift rather than the shift of interest (i.e., the presence of utensils).

\subsection*{Our Contributions}

In this work, we unify approaches to counterfactual generation under a common notion of a \textit{dataset interface}: a primitive that, given an input dataset and a user-specified shift, aims to return instances from the input distribution that exhibit the desired shift. We then study a number of existing strategies for implementing such an interface and find that ---  due to an undesirable mismatch between the distribution produced by the interface and that of the input dataset --- these approaches often introduce confounding shifts that can complicate model evaluation.

To mitigate this mismatch, we introduce a new implementation of a dataset interface that leverages Stable Diffusion together with Textual Inversion~\cite{gal2022image}. In particular, this implementation encodes each class in the input dataset as a token within the text-space of the diffusion model. By integrating these tokens into natural language prompts, our implementation can then generate counterfactual examples that conform to the input distribution while still exhibiting the desired shift. Overall, this implementation:
\begin{itemize}
    \item \textbf{Is tailored to the input dataset:} It can match key aspects of the original dataset, even for objects and attributes without a clear textual specification. For example, if the input dataset contains a specific breed of dog, our dataset interface can generate images matching that breed, even if the underlying diffusion model is unable to associate this breed with any natural language description.
    \item \textbf{Provides fine-grained control:} It can generate images of a target object with a significant degree of control over the desired distribution shift. This includes manipulating not only aspects such as backgrounds (e.g., ``on a beach'') and lighting (e.g., ``in studio lighting''), but also more fine-grained adjustments such as co-occurring objects (e.g., ``with a person'') and attributes of the objects themselves (e.g., ``lying down'').
    \item \textbf{Enables scalable counterfactual generation:} It is able to rapidly generate counterfactual examples, allowing us to evaluate a model's robustness across many possible failure modes and at scale.
\end{itemize}

Finally, leveraging this implementation, we create ImageNet$*$, a dataset interface for the ImageNet dataset~\cite{russakovsky2015imagenet}. We then demonstrate how we can use this interface to evaluate the performance of ImageNet-trained models under a diverse array of distribution shifts. In particular, due to our implementation's scalability and flexibility, we can use our interface to further a  \textit{shift-centric} perspective on model robustness, by categorizing how performance on \textit{different} types of shifts scales with model size, architecture, and pre-training regime.


%% file: sections/dataset_interfaces.tex
Let us return to our example of deploying a vision model for object classification. Suppose we wish to evaluate that model's ability to correctly classify images of a dog in a variety of contexts (e.g., ``on a beach''). To do so, we would like to collect \textit{counterfactual examples}, i.e., examples that exhibit the required distribution shift (``on a beach'') but still contain the original object (``dog'') as it appears in the training distribution. 

In order to unify such strategies for collecting counterfactual examples, we introduce the notion of a \textit{dataset interface}: a primitive that, given an input dataset and a user-specified shift, aims to return instances of a class from that dataset that exhibit the desired shift. In general, users do not have access to the original data-generating process, making it difficult to retrieve new examples with a specified context. A dataset interface thus serves as a proxy for the original dataset that enables users to control (and edit) the desired aspects of the surfaced images.

\paragraph{What makes a good dataset interface?} In order to facilitate wide-scale model evaluation, a dataset interface needs to fulfill three criteria: (1) images returned by the interface should exhibit the desired shift; (2) images returned by the interface should contain the specified object (as it appears in the input distribution); and (3) the interface needs to return images quickly in order to scale to wide-range model evaluation over many classes and contexts. In the following section, we discuss how existing approaches perform under these three criteria.


%% file: sections/dataset_interfaces_2.tex
\subsection{Existing Instantiations of Dataset Interfaces}
Currently, practitioners use a variety of techniques for surfacing counterfactual examples. Here, we discuss three categories of existing approaches --- manual data collection, text-to-image retrieval, and synthetic data generation --- that fit into the dataset interface framework. As a running example, let's return to our example of evaluating a ``dog'' classifier on a variety of backgrounds (e.g., ``a beach'').

\paragraph*{Manual data collection} 
\label{sec:manual_data_collection}
Perhaps the most straightforward strategy (as implemented in ObjectNet~\cite{barbu2019objectnet}, iWildCam \cite{beery2018recognition}) for acquiring counterfactual examples is to manually collect them from the real world. In our example, we could find 
the same types of dogs that appeared in our original dataset, and then take photos of those dogs on different backgrounds. This kind of data collection gives practitioners strong control over what the images they collect contain, but can be very expensive and time-consuming (and, for rarer objects like ``polar bear'', may be infeasible). 

\paragraph*{Text-to-image retrieval} A more common (and far more scalable) method is to find images from the internet using a textual query that matches our desired context (e.g., ``a photo of a dog on a beach''). One such approach (as implemented in ImageNet-R~\cite{hendrycks2020faces}, ImageNet-Sketch~\cite{wang2019learning}) is to query an image search engine like Bing or Google. Another strategy (as implemented in ADAVISION~\cite{gao2022adaptive}) is to retrieve images from a (huge) dataset of text-image pairs (e.g., LAION-5B~\cite{schuhmann2022laion}). For example, \texttt{clip-retrieval}~\cite{beaumont2022clipretrieval} uses a KNN index on top of a CLIP~\cite{radford2021learning} latent space --- a joint language-image embedding space learned with a contrastive objective --- to return images corresponding to a textual prompt. However, such text-to-image retrieval methods typically assume that a valid counterfactual example exists within the provided pool of images. We find that for many uncommon examples these methods are often unable to find a valid counterfactual example that includes both the desired object and shift (see Figure \ref{fig:plate_retrieval}).

\paragraph*{Synthetically generated data} Synthetic generation provides a more flexible mechanism for generating images in rarer contexts. Until recently, synthetic counterfactual examples were generated either by pre-processing images to induce distribution shifts (e.g., ImageNet-C~\cite{hendrycks2019benchmarking}), or rendering scenes using a 3D simulator~\cite{HamdiMG2018, alcorn2019strike, HamdiG2019, ShuLQY2020, JainCJWLYCJS2020, leclerc20213db}. However, these methods often produce images that are not photorealistic, or require involved processing steps (e.g., collecting a 3D scan of an object). More recently, taking advantage of current progress in generative models, other works~\cite{kattakinda2022invariant, wiles2022discovering} employ off-the-shelf text-to-image models such as Stable Diffusion~\cite{rombach2022high}, DALL-E 2~\cite{ramesh2022hierarchical}, and Imagen~\cite{saharia2022photorealistic} to generate photorealistic images conditioned on a textual prompt. In our running example, we could prompt Stable Diffusion to generate dogs on different backgrounds using a corresponding textual query (such as ``a photo of a dog on a beach.'')

However, synthetic generation can often introduce biases that introduce confounding shifts. For example, suppose that the input dataset only contains certain breeds of dogs. In this case, the diffusion model's conception of ``dog'' could differ from the dogs that actually exist in the ``input'' dataset (e.g., a completely different set of breeds). As a result, simply employing prompts that use the class name might not be sufficient to faithfully match the distribution of that dataset. In fact, we find that for a number of classes in the ImageNet dataset, there is a visual discrepancy between images generated by Stable Diffusion using the class name (e.g., ``dog'') and the images in that dataset itself. For instance, as we show in Figure~\ref{fig:mismatch}, there might be a mismatch in the specific type or appearance of the object (e.g., a subspecies of wolf, or columns in arch bridges). How can we then generate images that faithfully correspond to the distribution of the input dataset?


%% file: sections/method.tex
\begin{figure}[t!]
    \centering
    \begin{minipage}[b]{.58\textwidth}
      \centering
      \hspace{2cm}
      \includegraphics[width=1\linewidth]{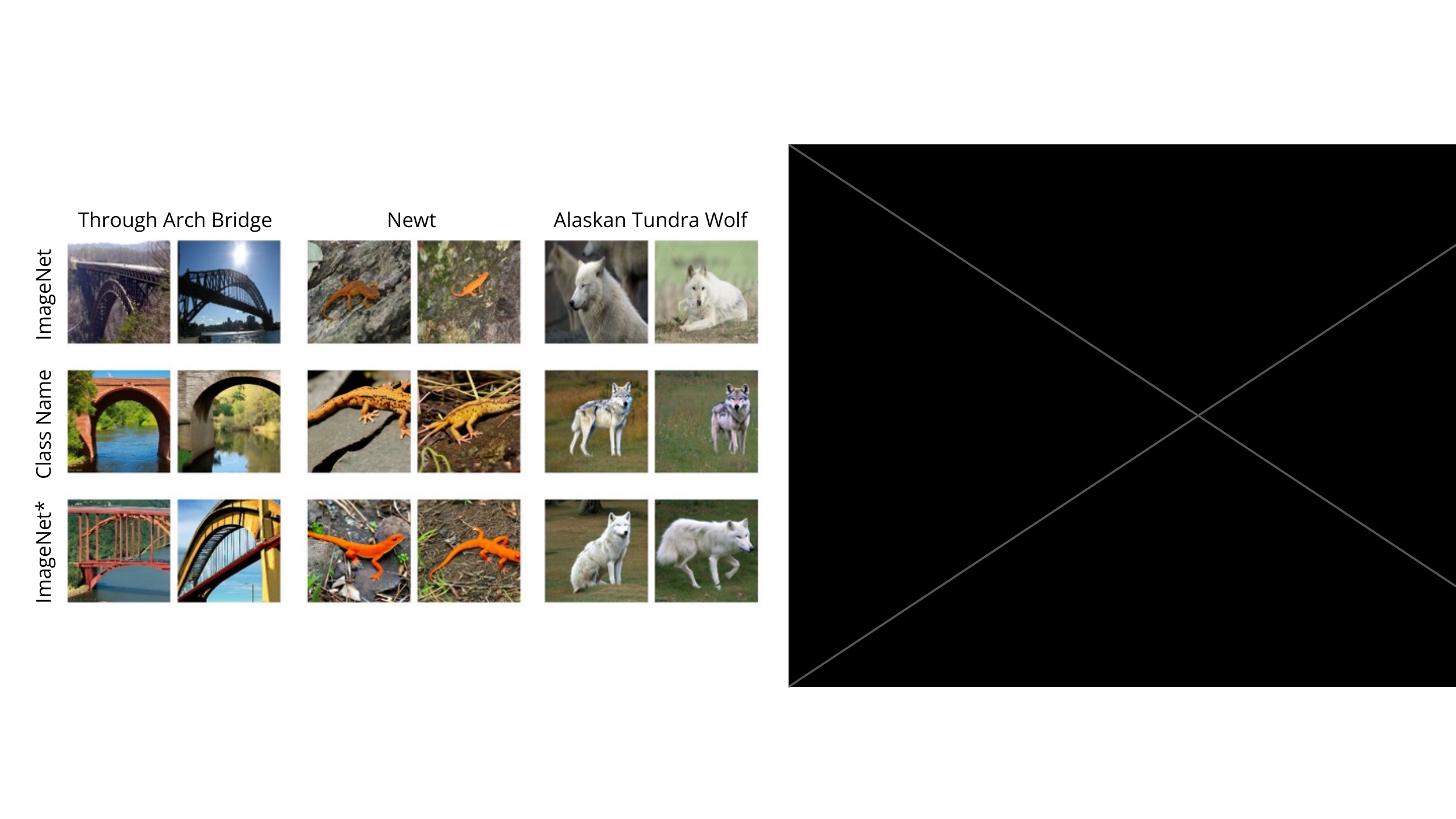}
      \caption{Examples of real images from ImageNet (\textbf{top}) and images generated  either by prompting Stable Diffusion with the class name (e.g., ``a photo of a newt'') (\textbf{middle}) or by using our dataset interface ImageNet$*$ (\textbf{bottom}). Note that for each class, the images from ImageNet$*$ match the original ImageNet distribution more closely than the images generated using the class name (see, e.g., the columns in the arch bridges). See Appendix~\ref{app:results} for further examples.}
      \label{fig:mismatch}
    \end{minipage}%
    \hfill
    \begin{minipage}[b]{.39\textwidth}
      \centering
      \includegraphics[width=1\linewidth]{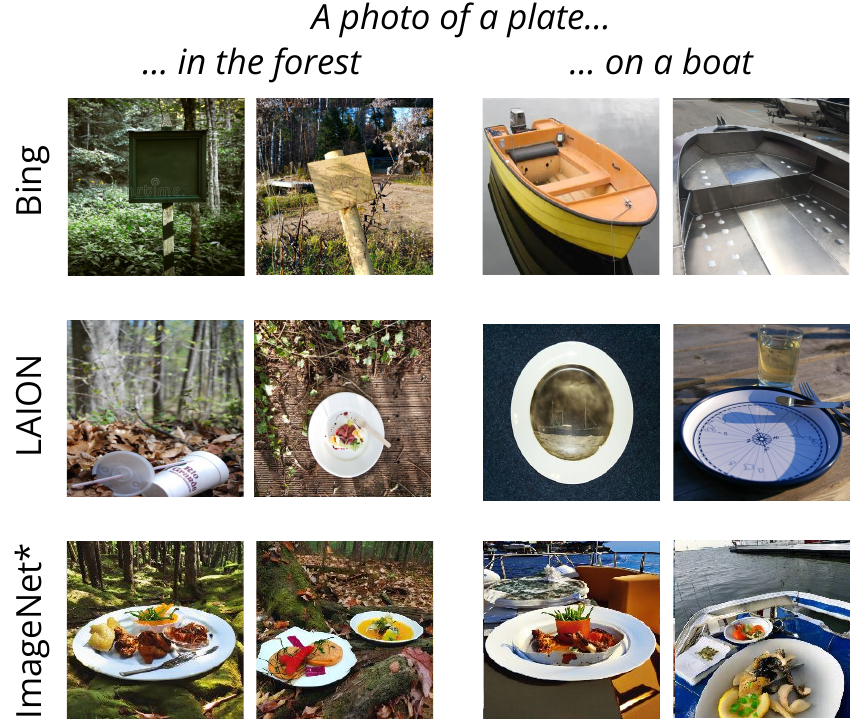}
      \caption{Examples of plates in various contexts collected from Bing (\textbf{top}), \texttt{clip-retrieval} (\textbf{middle}), or generated using our dataset interface ImageNet$*$ (\textbf{bottom}). Plates queried via text-to-image retrieval often miss either the object or shift, while the ImageNet$*$ plates contain both the object and shift.}
      \label{fig:plate_retrieval}
    \end{minipage}
\end{figure}

To overcome this mismatch,
we propose an implementation of a dataset interface that bridges the gap between the dataset interface and the corresponding input dataset. Specifically, we leverage recent work in \textit{personalized} text-to-image generation, which tries to incorporate user-provided visual concepts within a text-to-image diffusion model. By doing so, we can generate images that faithfully capture the properties of the corresponding class in the input dataset and avoid confounding shifts. In this work, we use Textual Inversion~\cite{gal2022image} (although it is possible to implement a similar interface with other personalized generation methods such as DreamBooth~\cite{ruiz2022dreambooth}).

\paragraph{Textual Inversion} Given a set of user-provided images containing a desired visual concept, Textual Inversion aims to find a ``word'' (token) $S_*$ in the diffusion model's text space to precisely capture that concept. This token can then be included in natural language prompts to generate images incorporating this desired concept. So, for example, using the prompt ``a monochrome photo of a $S_*$'' should result in generating a black and white image with the corresponding concept.

In order to create such a customized token $S_*$, Textual Inversion learns a corresponding embedding vector $v_*$ in the text embedding space of the diffusion model. To learn this embedding vector $v_*$, Textual Inversion freezes the weights of the entire pre-trained diffusion model and then finds $v_*$ that minimizes the diffusion model's original training objective, while using only the user-provided images that capture the desired visual concept paired with prompts containing $S_*$ (e.g., ``a photo of a $S_*$'').

\begin{figure*}[t!]
    \centering 
    \includegraphics[width=1\textwidth]{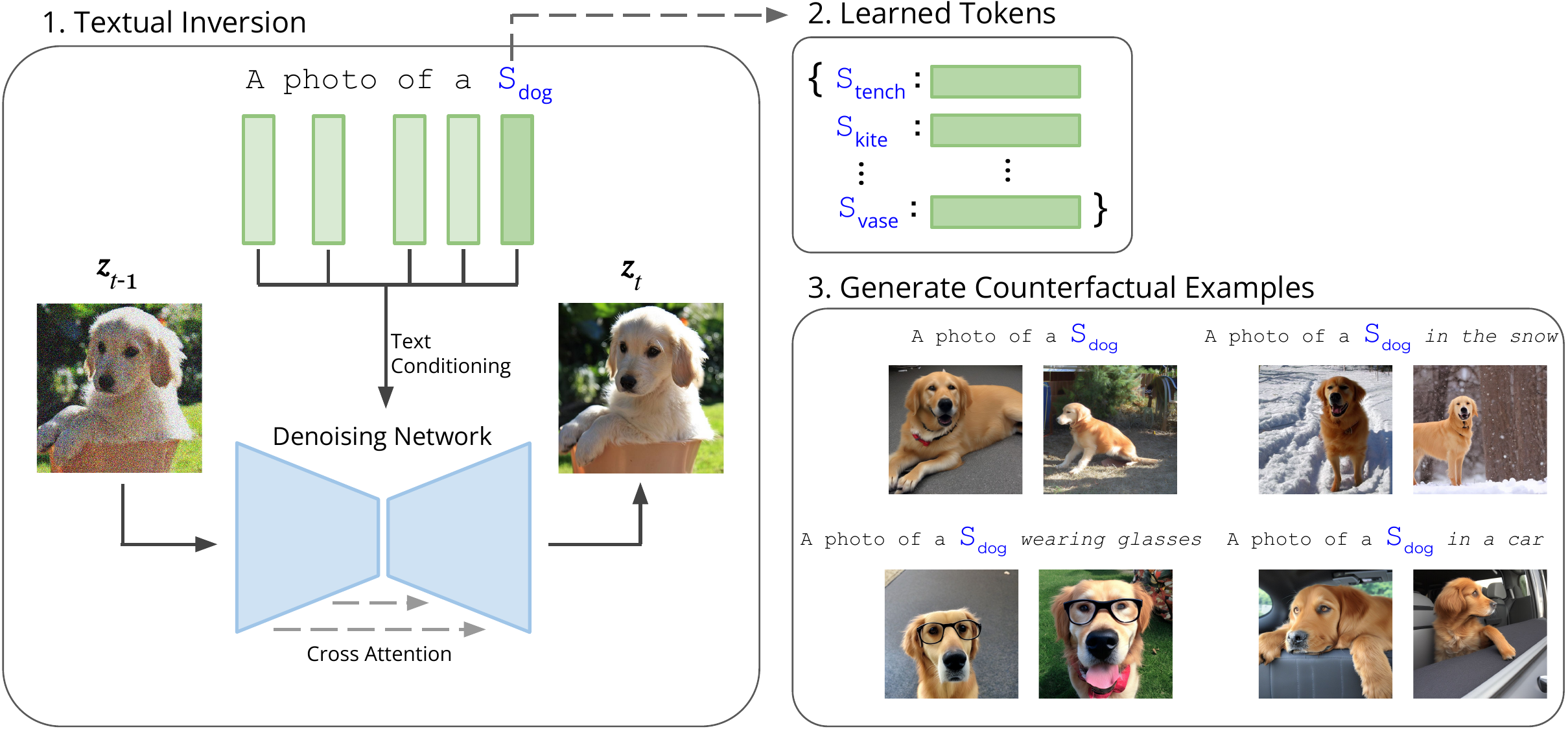}
    \caption{\textit{Construction of our dataset interface.} For each class in the input dataset, we use Textual Inversion (see Section~\ref{sec:method}) to learn a token in the text space of a text-to-image diffusion model. This token is intended to capture the distribution of the corresponding class. Then, by incorporating these tokens in natural language prompts, we can scalably generate a dataset of counterfactual examples.}
    \label{fig:method}
\end{figure*}

\paragraph{Encoding the input dataset as tokens in text space} 
With Textual Inversion in hand, we aim to guide a text-to-image diffusion model to generate images more closely aligned with the objects in the input dataset.
Specifically, for each class $c$ from that dataset, we run Textual Inversion on the training images of that class to learn an embedding vector $v_c$ for a corresponding new class token $S_c$. We can then incorporate these class tokens into our prompts to generate images under our desired shift. For example, to generate an image of a dog on the beach, we can use the prompt ``A photo of a $S_{dog}$ on the beach.'' We present an overview of our construction in Figure~\ref{fig:method}.

\subsection{Controlling the quality of counterfactual examples} \label{sec:threshold}
Text-to-image diffusion models can sometimes make mistakes, and as a result some of the counterfactual examples generated by our text prompts might either (1) not contain the original object or (2) not depict the desired distribution shift. We thus leverage CLIP \cite{radford2021learning} --- a large pre-trained model with a shared language-image embedding space --- to control the quality of ``candidate'' counterfactual examples based on these two criteria. Specifically, given a text label \textit{\textless class\textgreater } of a class (e.g., ``dog'') and a text description \textit{\textless shift\textgreater} of the desired distribution shift (e.g., ``on the beach''), we construct the following captions:
\begin{align*}
c_{class} = \textit{a photo of a \text{\textless class\textgreater}} && c_{shift} = \textit{a photo \text{\textless shift\textgreater}}
\end{align*}

Then, to quantify the presence of the original object and the desired distribution shift within an image, we measure the \textit{CLIP similarity}, i.e., the similarity between the CLIP embedding of the image and the text embeddings of captions $c_{class}$ and $c_{shift}$ respectively\footnote{For shifts describing art styles (e.g., ``a sketch of a ...'') we instead use $c_{class}$ = \textit{``a \text{\textless class\textgreater}''} as they are no longer a ``photo.''}. 
We use these metrics to automatically filter and remove images that do not meet the above criteria (see Appendix~\ref{app:threshold_selection} for details, and Appendix~\ref{app:clip-evalution} for yield rates.) A user study in Section~\ref{sec:distribution} confirms that this filtering step indeed improves the quality of the resulting dataset of counterfactual examples.



\subsection{ImageNet$*$} \label{sec:ImageNet*}
\begin{figure*}
    \centering 
    \includegraphics[width=1\textwidth]{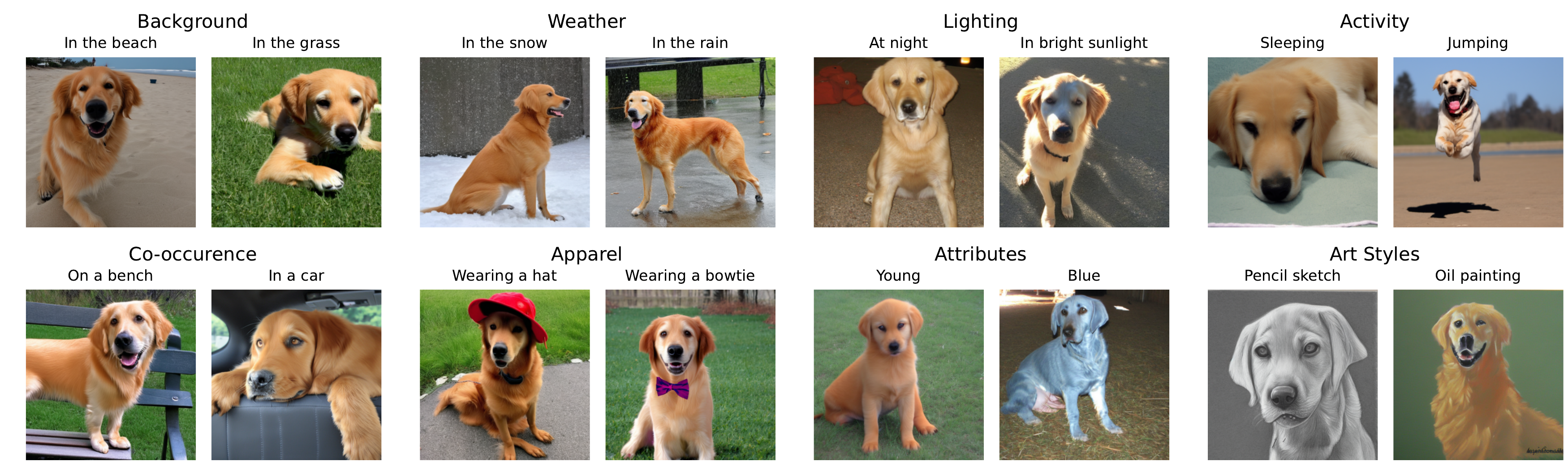}
    \caption{Examples of images generated using our dataset interface ImageNet$*$ for the golden retriever class and a variety of shifts.}
    \label{fig:dogs}
\end{figure*}

We apply the construction described above to create ImageNet$*$, a dataset interface for the ImageNet dataset (we defer results for other datasets to Appendix \ref{app:datasets}) and we publicly release the resulting set of 1,000 learned class tokens\footnote{\href{https://huggingface.co/datasets/madrylab/imagenet-star-tokens}{https://huggingface.co/datasets/madrylab/imagenet-star-tokens}}. Examples of images generated by ImageNet$*$ can be found in Figure~\ref{fig:dogs}.

In Section \ref{sec:benchmark}, we will use ImageNet$*$ to create a distribution shift robustness benchmark consisting of counterfactual examples for 23 different distribution shifts, including shifts in background, lighting, style, and object co-occurrence. We publicly release this benchmark as well\footnote{\href{https://huggingface.co/datasets/madrylab/imagenet-star}{https://huggingface.co/datasets/madrylab/imagenet-star}}, but we also encourage users to generate their own counterfactual examples tailored to their specific needs.

%% file: sections/distributions.tex
Having constructed ImageNet$*$ (see Section \ref{sec:ImageNet*}), we now evaluate the quality of our generated images.
Specifically, we use ImageNet$*$ to synthesize images of distribution shifts from five different categories --- ``at night'', ``blue'', ``in the beach'',``in the snow'', and ``sketch'' --- as well as ``base'' images which do not correspond to a specific shift (generated with a prompt ``a photo of a S*'').

As a baseline, we consider downloading images returned by the Bing search engine as well as retrieving them with \texttt{clip-retrieval}, an open source tool for scraping LAION, when queried with a natural language prompt containing the ImageNet class name. (See Appendix~\ref{app:bing} for details.)

We evaluate the quality of these images through a user study on the Amazon Mechanical Turk (MTurk) crowd-sourcing platform.  In this study, we show workers a grid of images, sampled from ImageNet$*$, scraped from the Bing engine, or retrieved from LAION-5B with \texttt{clip-retrieval}, with additional images from ImageNet as a control. We then ask the workers to identify which images contain (a) the target ImageNet class (e.g., ``golden retriever") and (b) the desired distribution shift (e.g., ``on the beach''). See Appendix~\ref{app:user_study} for further details. 

\begin{figure*}
    \centering 
    \includegraphics[width=1\textwidth]{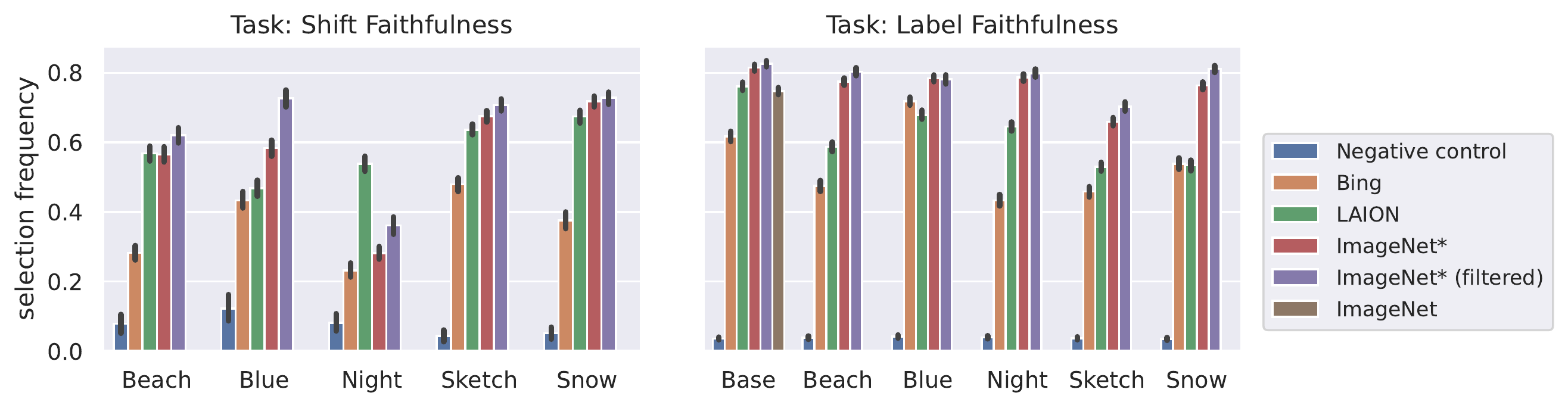}
    \caption{Selection frequencies (see Section \ref{sec:distribution}) for images from ImageNet, images scraped using Bing, image retrieval from LAION, and images generated with ImageNet$*$ when asking workers to identify either presence of a specific distribution shift (\textbf{left}) or presence of a ImageNet class (\textbf{right}) in the image. The ImageNet$*$ images exhibit the desired context and object more often than those from both Bing and LAION. Filtering the images with the CLIP metrics consistently increases selection frequency for both the shift and the object.}
    \label{fig:user_study}
\end{figure*}

In Figure~\ref{fig:user_study}, we report the \textit{selection frequency}, i.e., the fraction of images selected by the workers, for each of these two tasks. We find that according to the workers, the images generated by our framework exhibit the desired shift and object of interest more often than images scraped using Bing across each distribution shift. Also, querying LAION with \texttt{clip-retrieval} turns out to be competitive with ImageNet$*$ in retrieving images with the desired shift, but these images more often do not contain the desired object. Finally, we observe that filtering the generated counterfactual examples with our filtering scheme (see \ref{sec:threshold}) improves the selection frequency for both tasks.

%% file: sections/debugging.tex
\begin{figure*}
    \centering 
    \includegraphics[width=1\textwidth]{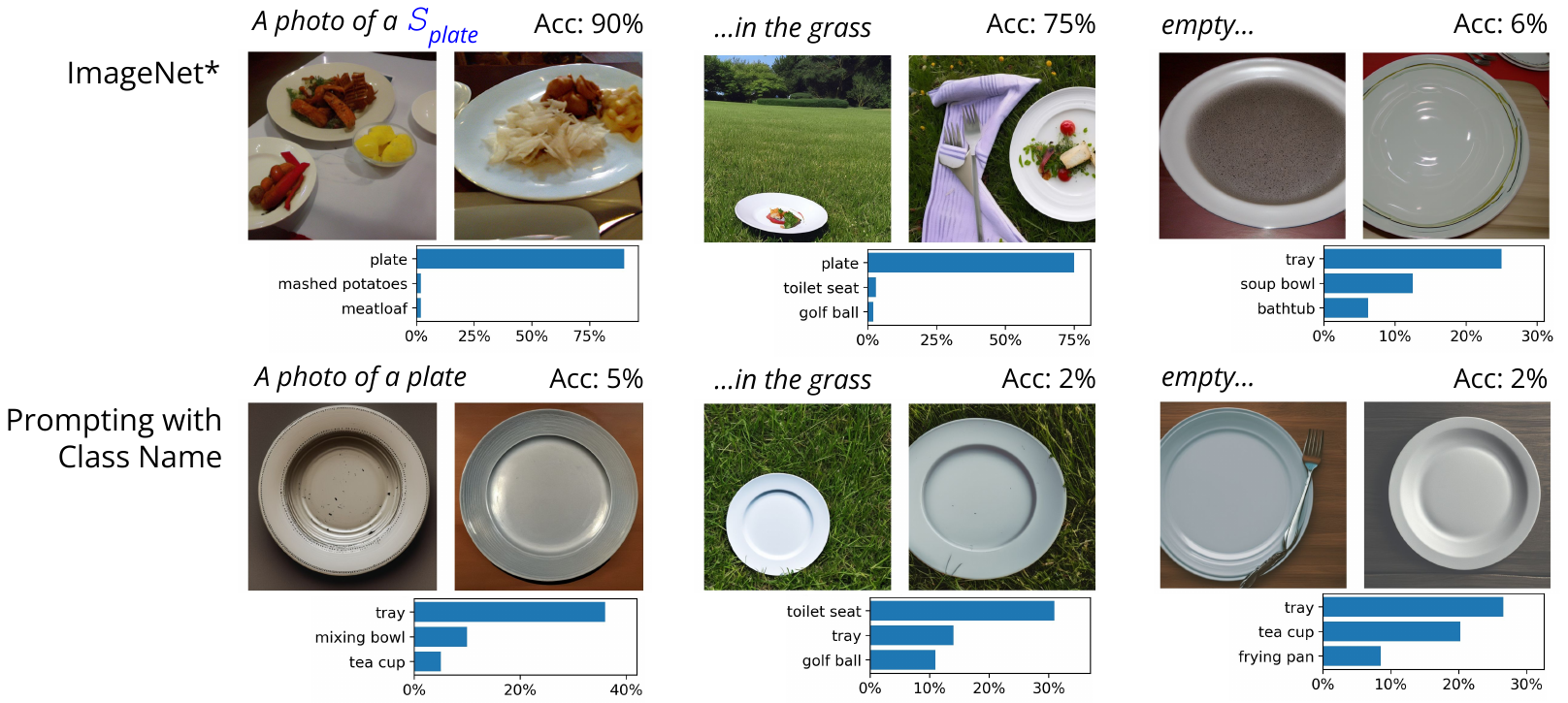}
    \caption{Images of plates generated by our dataset interface ImageNet$*$ (\textbf{top}) and by prompting Stable Diffusion with the class name ``plate'' (\textbf{bottom}), alongside the top predicted classes of an ImageNet-trained ResNet50 on these images. While generating ImageNet$*$ examples ``in the grass'' causes only a slight drop in accuracy, prompting Stable Diffusion with \textit{``a plate in the grass''} degrades the model's accuracy to 2\% due to the additional confounding factor of emptiness. 
    }
    \label{fig:debugging}
\end{figure*}

Capturing certain class-specific model failures may require executing fine-grained adjustments in particular scenarios (such as adding a harness on a dog or putting a fish in a tank). Our dataset interface provides exactly this kind of debugging capability while avoiding the unintended effect of secondary ``confounding'' shifts. To illustrate this, let us return to our example task of deploying an ImageNet-trained model. Suppose that we would like to examine the model's performance on images of plates in the grass. When prompting Stable Diffusion with a query that uses the class name (``a photo of a plate in the grass''), we find that our classifier achieves an accuracy of \textit{only 2\%} on the resulting images! Is the grassy background really such a catastrophic failure mode for our model? 

If, instead, we use ImageNet$*$ to generate counterfactual examples of ``plates on the grass", we find that our classifier's accuracy only slightly drops from 90\% to 75\% (see Figure~\ref{fig:debugging}).  What causes this discrepancy? It turns out that the ``failure case'' identified when prompting Stable Diffusion with natural language is actually an extreme example of a confounding shift. Indeed, recall that ImageNet plates usually have food on them (c.f., Figure~\ref{fig:bing_figure}). However, the plates generated by Stable Diffusion are almost exclusively empty, even when using a prompt that in principle does not introduce any shift (i.e., ``a photo of a plate'').

\begin{figure*}[t!]
    \centering 
    \includegraphics[width=0.9\textwidth]{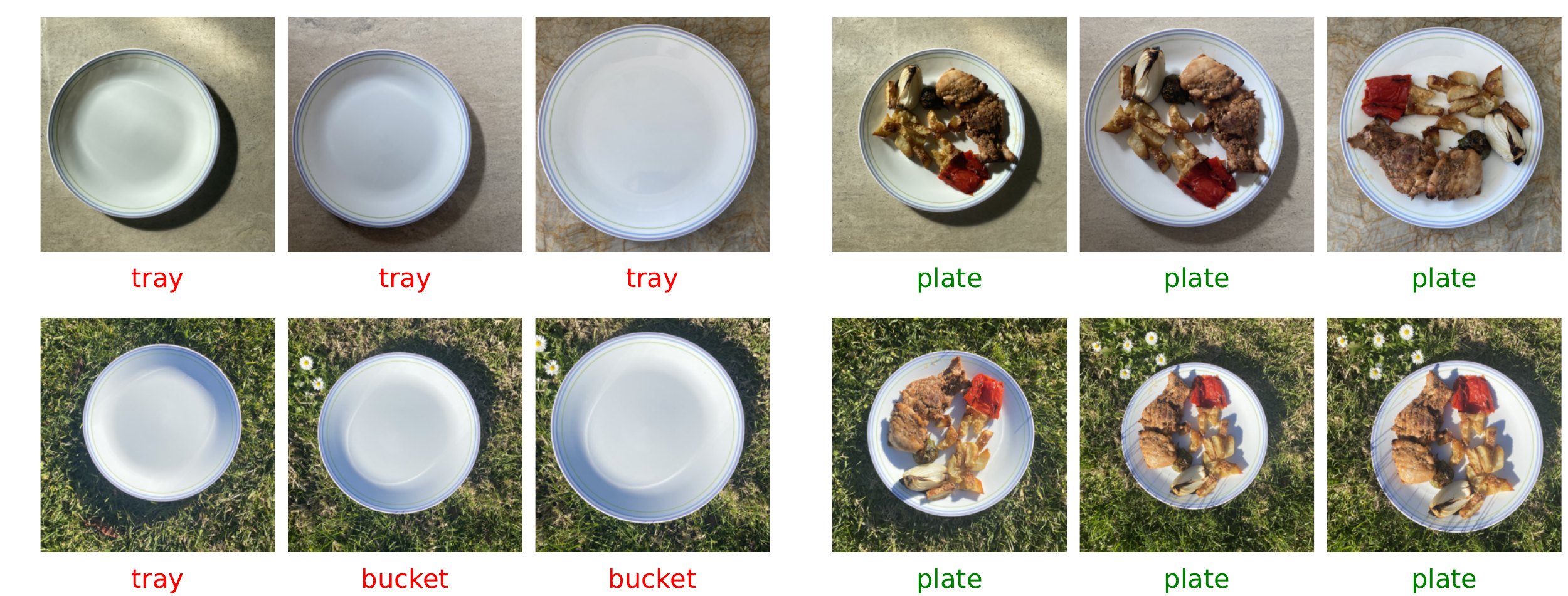}
    \caption{Real images of plates, with and without food and either on a table or in the grass. Below each image is the predicted class by an ImageNet-trained ResNet50.}
    \label{fig:real_images}
\end{figure*}

To assess to what degree this confounding shift of ``emptiness'' is detrimental for our ImageNet classifier, we use ImageNet$*$ to generate counterfactual examples of empty plates. We then find that the classifier's accuracy decreases to only 6\% (so, similar to the 2\% we observed before). To further confirm that the failure mode is indeed caused by emptiness and not the presence of grass, we took real photos of a plate in each of these contexts and evaluated our model on them (see Figure~\ref{fig:real_images}).

So, as we have seen, dataset interfaces enable us to test distribution shifts in \textit{isolation}, i.e., without introducing confounding shifts that can produce misleading results. In Appendix~\ref{app:results}, we discuss additional examples of using our interface for precise model debugging. 

%% file: sections/benchmark.tex
\begin{figure*}
    \centering 
    \includegraphics[width=1\textwidth]{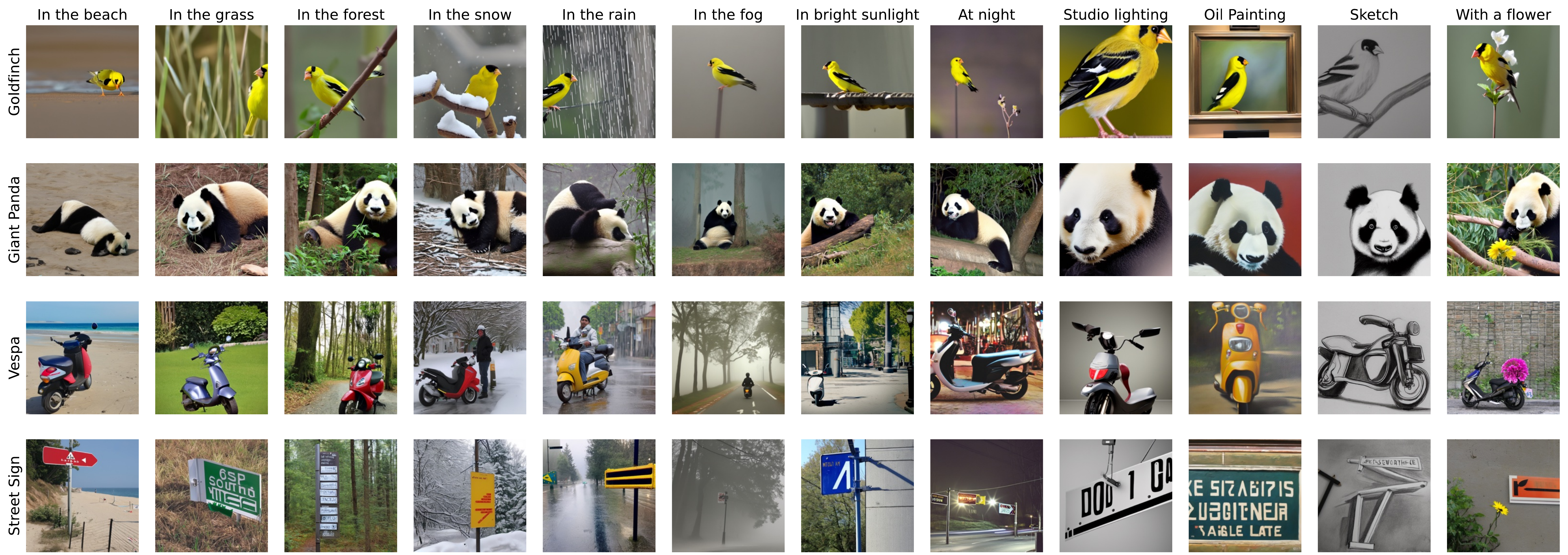}
    \caption{Examples of images generated with ImageNet$*$ for a variety of distribution shifts. These images are a subset of the benchmark described in Section~\ref{sec:benchmark}. See Appendix~\ref{app:results} for further examples.}
    \label{fig:benchmark}
\end{figure*}

Our framework's scalability enables us to rapidly assess a model's performance on a wide variety of distribution shifts. As a result, we can take a \textit{shift-centric} perspective on robustness; that is, we can evaluate models on many types of distribution shifts at once, and then categorize variations in these models' behavior.

\paragraph{A benchmark for distribution shift robustness.}
Using our dataset interface ImageNet$*$, we generate images for 23 shifts, including changes in background, weather, lighting, style, attributes, and co-occurrence (see Figure~\ref{fig:benchmark} for examples, and Appendix \ref{app:benchmark_setup} for a full list). We then evaluate a variety of image classification models varying architectures, training regimes, pre-training schemes, and input resolutions.

\begin{figure}[t!]
    \centering 
    \includegraphics[width=0.9\columnwidth]{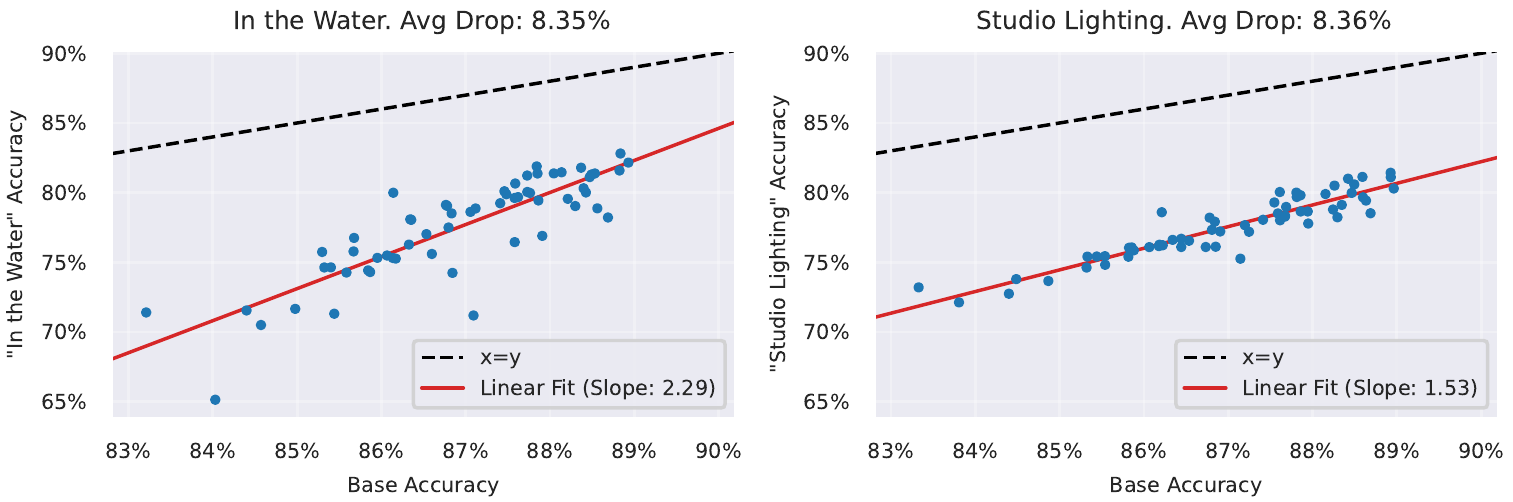}
    \caption{
        Accuracy on images generated with a distribution shift vs. accuracy on images generated with the base prompt (i.e., ``a photo of $S_*$'') for two distribution shifts over a sweep of ImageNet models.}
    \label{fig:on_the_line}
\end{figure}

We can now categorize the behavior of each shift according to two criteria. The first criteria, the shift's \textit{absolute impact}, encapsulates the shift's overall severity, and can be measured as the average difference between the models' performance on the base generated images and the corresponding counterfactual examples. The second criteria, the \textit{ID/OOD slope} captures the degree to which improving model accuracy on in-distribution images also boosts its performance under the distribution shift. We measure this quantity by plotting the accuracy of each model on the base generated images versus on the counterfactual examples (as in~\cite{taori2020when,miller2021accuracy}), and then calculating the slope of the best-fit-line (see Figure \ref{fig:on_the_line} for two examples).

\begin{figure}[t!]
    \begin{subfigure}[b]{0.765\textwidth}
        \centering
        \includegraphics[width=1\linewidth]{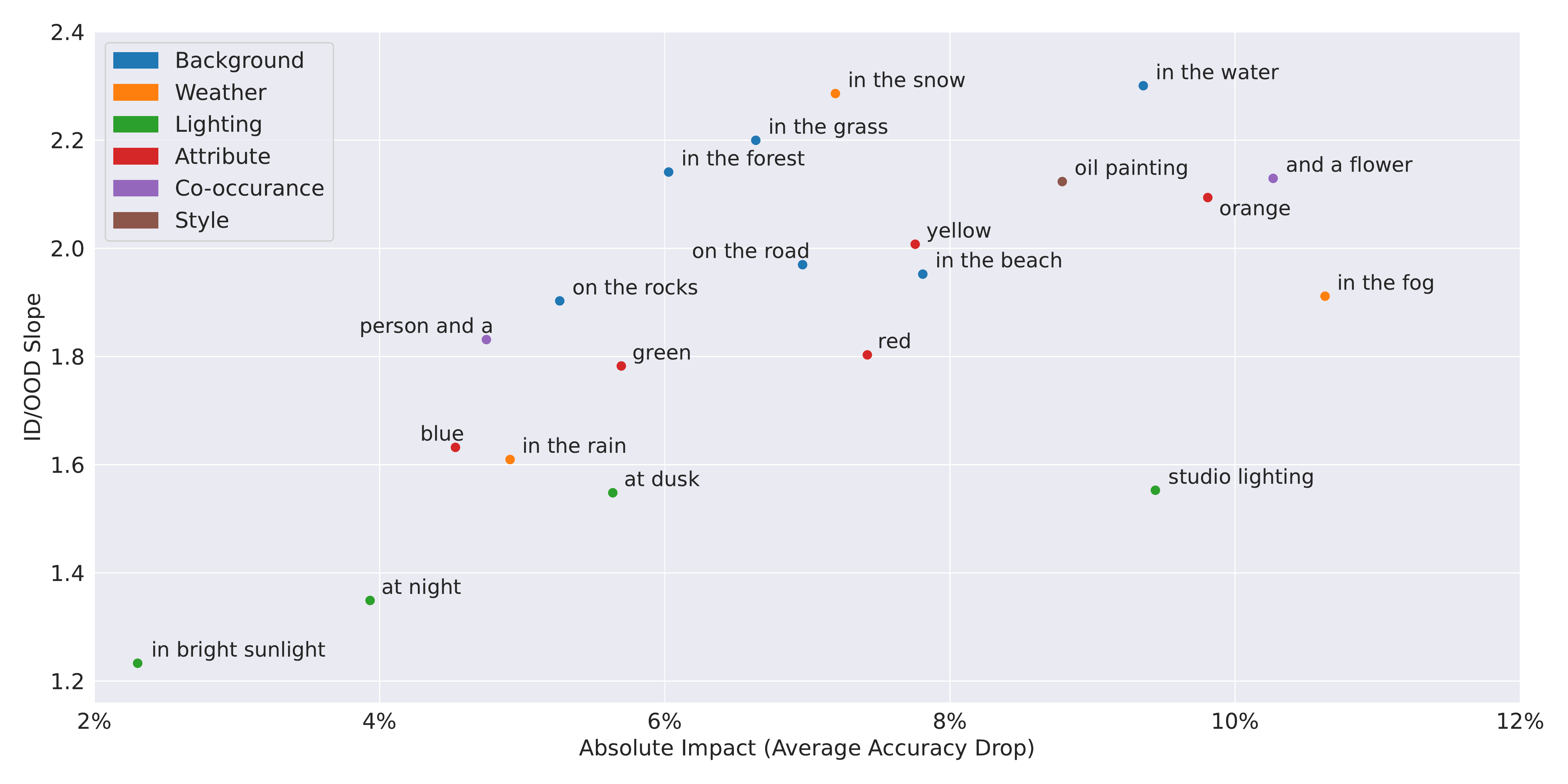}
    \end{subfigure}\hfill
    \begin{subfigure}[b]{0.234\textwidth}
        \centering
        \includegraphics[width=\linewidth]{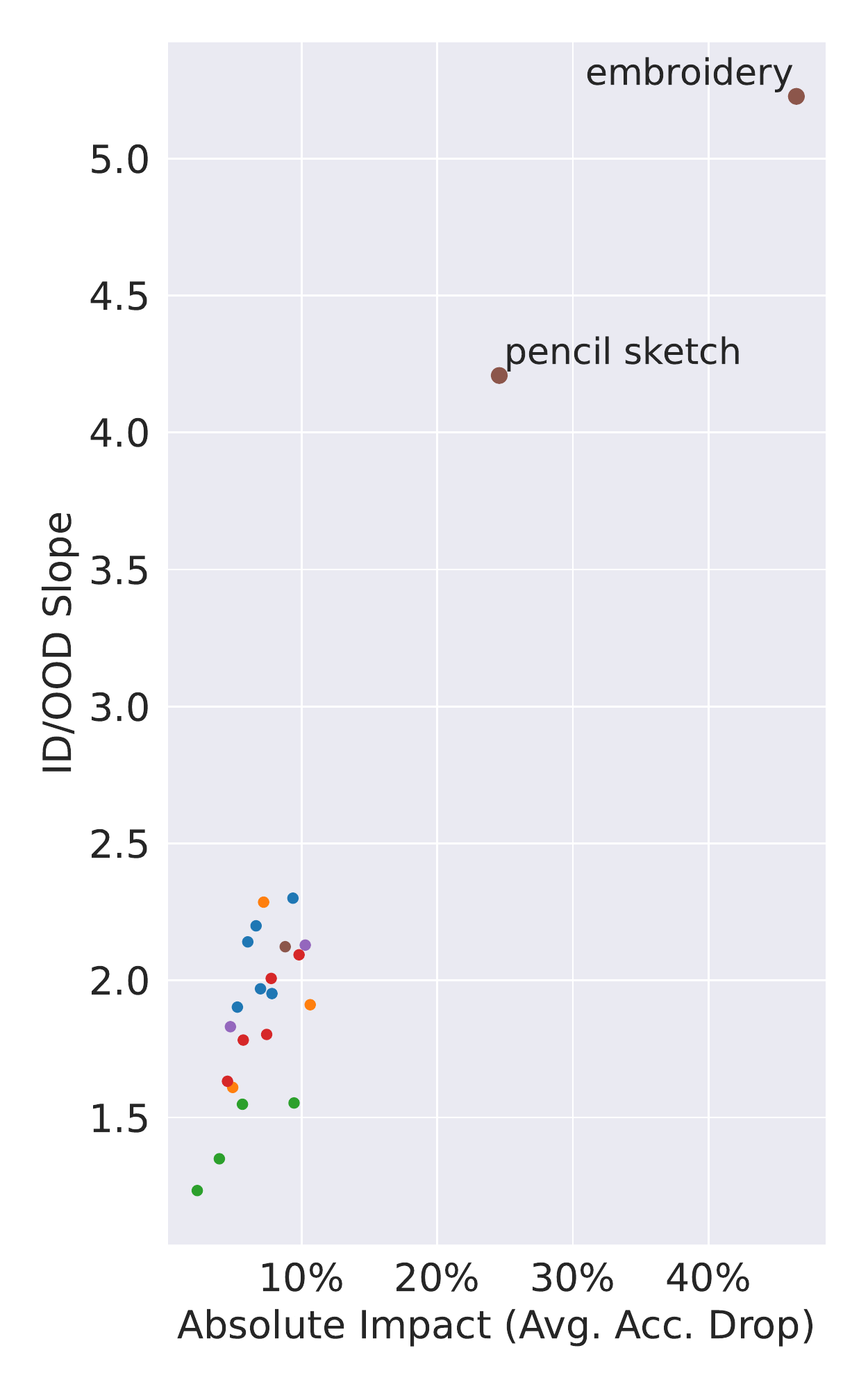}
    \end{subfigure}
    \caption{
        For each of the shifts in the benchmark, we plot the ID/OOD slope (degree to which in-distribution accuracy improves accuracy under the shift) versus absolute impact (average drop in accuracy due to the shift). The ``'pencil sketch'' and ``embroidery'' shifts are shown separately on the right, as their absolute impact and ID/OOD slope fall far from all other shifts (points for the other shifts are shown as reference). Note that changes in lighting (e.g., ``at dusk'') and attribute (e.g., ``red'') have low ID/OOD slope, indicating that the models' performance on these images remains static regardless of in-distribution accuracy.
        }
    \label{fig:evaluation}
\end{figure}

In Figure~\ref{fig:evaluation}, we plot the absolute impact and the ID/OOD slope for each one of the considered distribution shifts. We find that different types of shifts result in different scaling behaviors across these two criteria. For example, even though ``in the water'' and ``studio lighting'' have similar absolute impacts, ``in the water'' has a higher ID/OOD slope. Therefore, while boosting the in-distribution accuracy for ImageNet can help improve the model's performance on images ``in the water'', the model's performance on ``studio lighting'' is much more static. More broadly, we find that shifts based on lighting (e.g., ``studio lighting'')  have lower ID/OOD slope than shifts based on background (e.g., ``in the grass''), with attributes (e.g., ``red'') in between. 

%% file: sections/related_work.tex
\paragraph{Benchmarks for distribution shift robustness} Many robustness benchmarks evaluate model performance under specific distribution shifts by collecting real images. 
These include shifts in style~\cite{hendrycks2020faces, wang2019learning}, object pose~\cite{barbu2019objectnet}, background~\cite{beery2018recognition}, time/location~\cite{christie2018functional, hendrycks2020faces}, and data pipelines~\cite{recht2018imagenet}. Other works create synthetic distribution shift benchmarks, often by preprocessing the images of an ``in-distribution'' dataset to induce a shift. One common strategy here is to synthetically create shifts in background by pasting the foreground of the target image onto an alternate background image~\cite{xiao2020noise, sagawa2019distributionally, kattakinda2022invariant}. In particular, ImageNet-C \cite{hendrycks2019benchmarking} applies a set of transformations such as blur and synthetic fog on top of images to simulate real-world corruptions. Finally, TILO~\cite{lynch2022evaluating} uses Stable Diffusion to generate images of vehicles with variations in backgrounds and lighting.

\paragraph{Identification of failure modes through counterfactual examples} 
There are a number of works that aim to diagnose model failures by evaluating them on counterfactual examples. One line of such work leverages 3D rendering software to synthesize objects with varying geometry and object pose~\cite{HamdiMG2018, alcorn2019strike, HamdiG2019, ShuLQY2020, JainCJWLYCJS2020, leclerc20213db}. 
Of these, 3DB~\cite{leclerc20213db} is the closest to our work, as in addition to object pose, they allow control over aspects including lighting, background, texture, and object co-occurrence. However, their framework still requires the user to first acquire a 3D model of the object of interest.

On the other hand, ADAVISION \cite{gao2022adaptive} introduces an interactive process for identifying model failures by repeatedly querying for real images from LAION-5B 
and optimizing the query to more closely match the model's misclassifications. However, ADAVISION requires user intervention at each step to verify model failures.
\citet{wiles2022discovering}, in turn, propose a framework for automatically surfacing model failures by generating images
with a text-to-image generative model, clustering misclassified inputs, and then using a image-to-text model to caption these clusters.
Finally, \citet{jain2022distilling} synthesize prototypical examples of challenging subpopulations by automatically captioning model failure modes and then generating the images via Stable Diffusion. 

\paragraph{Personalized text-to-image generation} While our dataset interface utilizes textual inversion to learn personalized concepts, there have been many recent works in the context of personalized text-to-image generation. 
These approaches aim to incorporate user-provided visual concepts (e.g., an object or style) into text-to-image generation.
One family of techniques
uses a guiding image to further condition generation~\cite{jeanneret2022diffusion, yuan2022not, kattakinda2022invariant}. Specifically, D3S \cite{kattakinda2022invariant} first pastes the foreground of a given object onto a background and then uses the resulting image to guide Stable Diffusion. 
\citet{yuan2022not} generate images in a desired target domain by conditioning on an image from the source dataset and a prompt that describes the target domain.

Another approach to personalized generalization, taken by methods such as Textual Inversion~\cite{gal2022image} and DreamBooth~\cite{ruiz2022dreambooth}, allows users to directly encode a desired concept within the text space of the text-to-image model. While Textual Inversion learns a new token within a frozen text-to-image model, DreamBooth fine-tunes the full model. By allowing the original diffusion model weights to change, DreamBooth offers greater capability for personalization at the potential cost of degrading the generation of concepts already known to the model. 

%% file: sections/conclusion.tex
In this work, we introduce the notion of a dataset interface: a framework that, given an input dataset and user-specified shift, returns instances from that input distribution that exhibit the desired shift. While there are a number of ways to implement of such an interface, we find that they often introduce confounding shifts due to an undesirable mismatch between the interface and the input dataset. To mitigate this issue, we put forth an implementation of a dataset interface that leverages Textual Inversion to tailor counterfactual generation more closely to the input dataset. 
In addition to enabling fine-grained model debugging, our dataset interface implementation allows users to simultaneously evaluate a diverse array of distribution shifts, making it possible to take a more ``shift-centric'' perspective on model robustness.

There are several avenues for further investigation. While our dataset interface implementation leverages natural language descriptions to represent distribution shifts, one could instead attempt to automatically learn a representation of a shift using user-provided example images. Another potential direction to explore is to use counterfactual examples generated by such an interface to improve a model's robustness (e.g., by incorporating the generated images into the training pipeline).

%% file: sections/acks.tex
Work supported in part by the NSF grants CNS-1815221 and DMS-2134108, and Open Philanthropy. This material is based upon work supported by the Defense Advanced Research Projects Agency (DARPA) under Contract No. HR001120C0015. SJ is supported by the Two Sigma Diversity Fellowship. 

Research was sponsored by the United States Air Force Research Laboratory and the United States Air Force Artificial Intelligence Accelerator and was accomplished under Cooperative Agreement Number FA8750-19-2-1000. The views and conclusions contained in this document are those of the authors and should not be interpreted as representing the official policies, either expressed or implied, of the United States Air Force or the U.S. Government. The U.S. Government is authorized to reproduce and distribute reprints for Government purposes notwithstanding any copyright notation herein.

%% file: sections/appendix.tex
\section{Setup Details}
\label{app:setup_details}
\input{sections/appendix/setup_details}

\clearpage
\section{Additional Results and Visualizations}
\label{app:results}
\input{sections/appendix/results}

\clearpage
\section{CLIP Metrics}
\label{app:clip-evalution}
\input{sections/appendix/clip}

\clearpage
\section{Experiments on Additional Datasets}
\label{app:datasets}
\input{sections/appendix/datasets}

%% file: sections/appendix/setup_details.tex
\subsection{Textual Inversion}
To learn each token, we run textual inversion with $3,000$ optimization steps. We use an adam optimizer wth a constant learning rate schedule, a learning rate of $5e-4$, $\beta_1=0.9$, $\beta_1=0.999$, and weight decay $1e-2$. Our hyperparameters follow the HuggingFace textual inversion script at:

\href{https://github.com/huggingface/diffusers/tree/main/examples/textual_inversion}{https://github.com/huggingface/diffusers/tree/main/examples/textual\_inversion}.

\subsection{Scraping images from Bing} \label{app:bing}
For the Bing engine baseline, we query Bing with natural language prompts. We leverage the scraping library \href{https://pypi.org/project/bing-image-downloader/}{bing-image-downloader}, and scrape the top 50 images per class. 

\subsection{Setting CLIP Thresholds} \label{app:threshold_selection}
Here we describe our procedure for setting thresholds our CLIP similarity metric when filtering.

To set the similarity threshold for the presence of the object in a generated image, we evaluate the similarity between the embedding of $c_{class}$ and every image of that class in the ImageNet validation set. We set then the threshold at the $20^{th}$ percentile of the CLIP similarities.

To set the similarity threshold for the presence of the distribution shift in a generated image, we first evaluate the CLIP similarity between the embedding of $c_{shift}$ and every generated image in that distribution shift. For each of a fixed set of percentile values, we visually inspect a small number of images with similarities around that percentile, and select as our threshold the lowest percentile at which all inspected images exhibit the desired distribution shift.

\subsection{User Study} \label{app:user_study} We verify that our generated
counterfactual examples for the ImageNet dataset contain the desired
distribution shift and the object of interest through a user
study on the Amazon Mechanical Turn (MTurk) crowd-sourcing platform. Below we
describe the procedure of our study.

\begin{figure}
    \centering
    \includegraphics[width=0.9\textwidth]{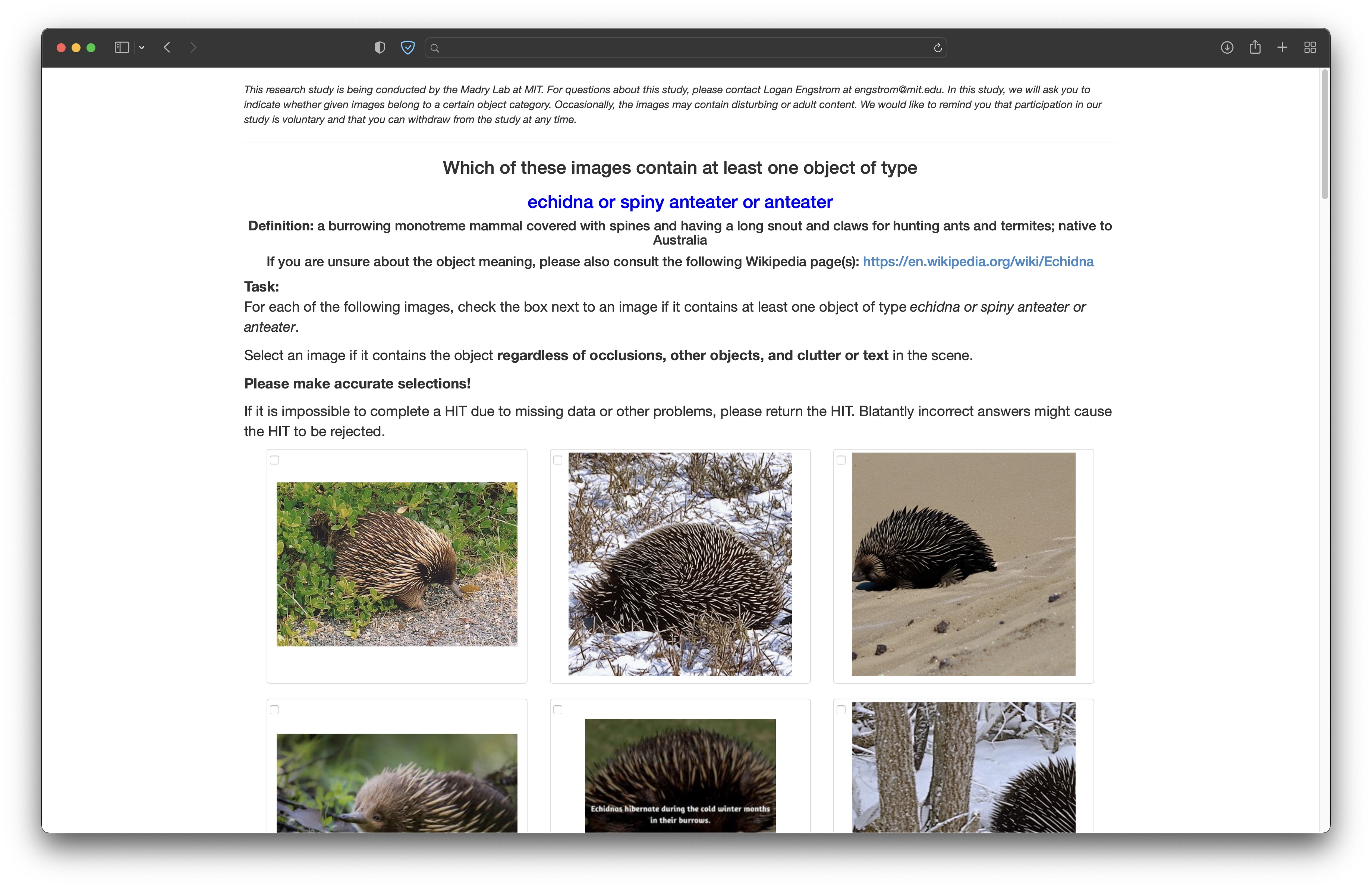}
    \caption{User study: label task. In this task we ask crowd-workers to verify
    that the generated images correspond to the desired label.}
    \label{appfig:label}
\end{figure}

\begin{figure}
    \centering
    \includegraphics[width=0.9\textwidth]{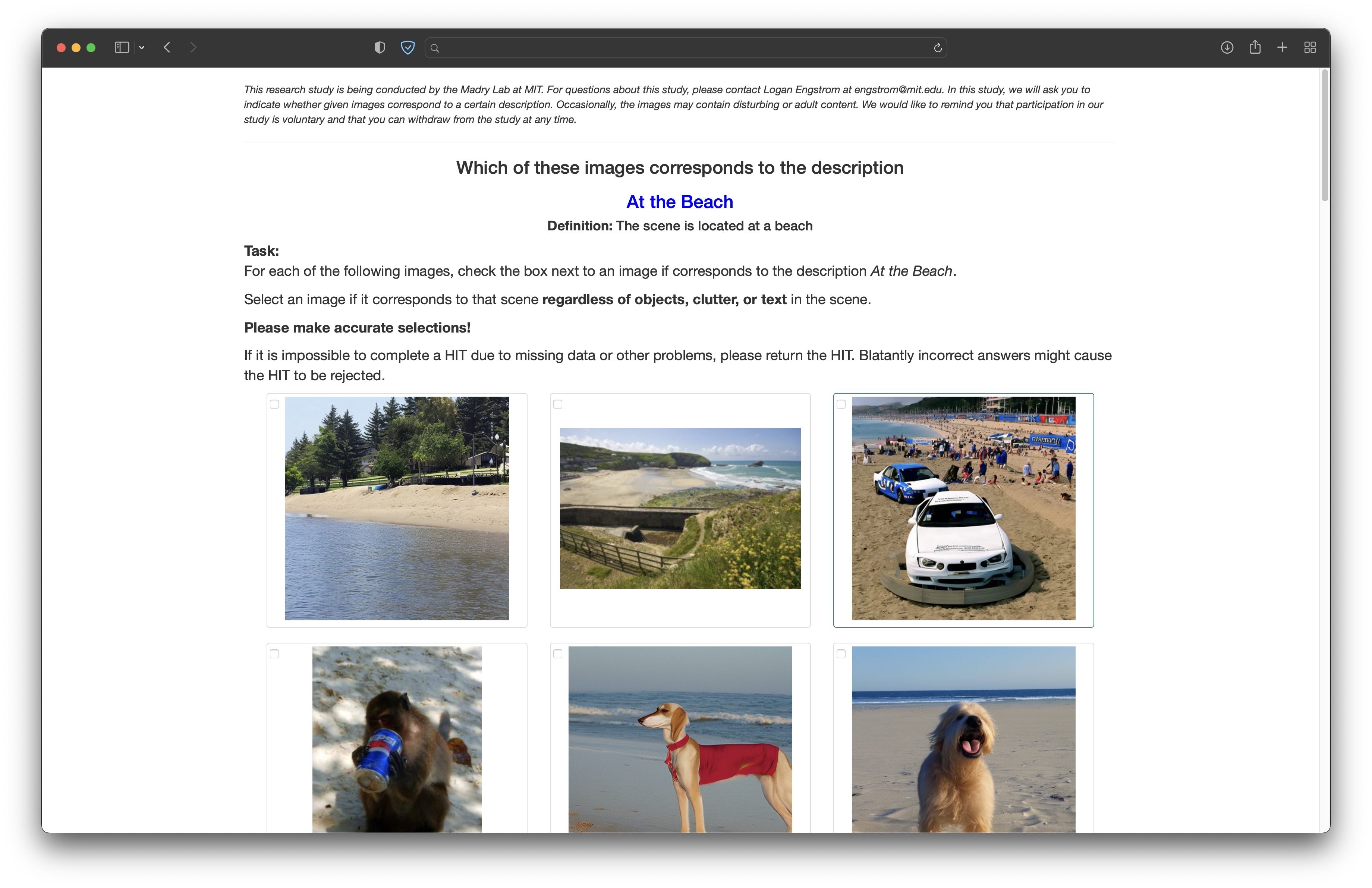}
    \caption{User study: distribution shift task. In this task we ask
    crowd-workers to verify that the generated images correspond to the desired
    distribution shift.}
    \label{appfig:shift}
\end{figure}

\paragraph{Procedure} We send grids of 48 images to Amazon Mechanical Turk
workers to label (pictured in Figure~\ref{appfig:label} and
Figure~\ref{appfig:shift}). Each grid contains a single label, and the workers
are asked to label all the instances of this label. This label depends on the
task; for the "label verification" task this label is an ImageNet label, and for the
"shift verification" this label is a distribution shift. We send each grid to 5
workers to label. We then measure the \textit{selection frequency} for every
image in the grid: the frequency at which workers selected the image as
corresponding to the given label in the grid. We employ the selection frequency
as a proxy score for how likely it is for the label to truly correspond to a
given image.

\clearpage
\subsection{Evaluating Distribution Shift Robustness} \label
{app:benchmark_setup}

We generate 50000 images in 23 different shifts, listed in Table \ref{tab:thresholds} along with the threshold we use to filter each shift. We evaluate the images on a sweep of image classification models obtained from the \texttt{timm} model repository \cite{rw2019timm}. We list all models in the file \textit{models.txt} provided with the code.

Applying our CLIP metrics to filter the images results in an uneven number of images per class. In fact, for some of our more difficult shifts a class might have no images that pass the filter  threshold. When calculating the accuracy of a model for images in a given shift, we only consider classes with at least five images remaining after filtering. Among these classes we measure the average per-class accuracy, and correspondingly measure the average per-class accuracy of the same classes among the base generated images. The difference between these two numbers is the accuracy drop we record for that model.

In Table \ref{tab:thresholds} we list each of the distribution shifts we include in our benchmark along with the base prompt that we use to represent in-distribution samples. For each shift we list the full prompt that is input to Stable Diffusion, the CLIP threshold we set for filtering the generated images of that shift, and the \% yield for images remaining after filtering with both metrics (for both presence of the desired distribution shift and object of interest).

\begin{table}[h!]
    \centering
    \begin{tabular}{l|l|c|c}
        Shift Name & Prompt & CLIP Threshold & \% Yield \\
        \midrule
        base & A photo of a \textless class\textgreater & $-$ & 92.6\% \\
        ``in the grass'' & A photo of a \textless class\textgreater in the grass & 0.127 & 80.3\% \\
``in the beach'' & A photo of a \textless class\textgreater in the beach & 0.175 & 62.9\% \\
``in the forest'' & A photo of a \textless class\textgreater in the forest & 0.153 & 67.3\% \\
``in the water'' & A photo of a \textless class\textgreater in the water & 0.163 & 60.1\% \\
``on the road'' & A photo of a \textless class\textgreater in the road & 0.154 & 64.5\% \\
``on the rocks'' & A photo of a \textless class\textgreater in the rocks & 0.124 & 76.6\% \\
``in the snow'' & A photo of a \textless class\textgreater in the snow & 0.160 & 73.2\% \\
``in the rain'' & A photo of a \textless class\textgreater in the rain & 0.173 & 48.3\% \\
``in the fog'' & A photo of a \textless class\textgreater in the fog & 0.152 & 59.3\% \\
``in bright sunlight'' & A photo of a \textless class\textgreater in bright sunlight & 0.124 & 89.9\% \\
``at dusk'' & A photo of a \textless class\textgreater at dusk & 0.158 & 61.9\% \\
``at night'' & A photo of a \textless class\textgreater at night & 0.147 & 61.1\% \\
``studio lighting'' & A photo of a \textless class\textgreater in studio lighting & 0.140 & 66.6\% \\
``blue'' & A photo of a blue \textless class\textgreater & 0.163 & 59.1\% \\
``green'' & A photo of a green \textless class\textgreater & 0.190 & 51.3\% \\
``red'' & A photo of a red \textless class\textgreater & 0.167 & 59.6\% \\
``yellow'' & A photo of a yellow \textless class\textgreater & 0.212 & 43.3\% \\
``orange'' & A photo of a orange \textless class\textgreater & 0.216 & 41.0\% \\
``person and a'' & A photo of a person and a \textless class\textgreater & 0.181 & 29.9\% \\
``and a flower'' & A photo of a \textless class\textgreater and a flower & 0.148 & 61.9\% \\
``oil painting'' & An oil panting of a \textless class\textgreater & 0.214 & 67.2\% \\
``pencil sketch'' & A black and white pencil sketch of a \textless class\textgreater & 0.223 & 61.8\% \\
``embroidery'' & An embroidery of a \textless class\textgreater & 0.259 & 33.0\% \\
    \end{tabular}
    \caption{Full prompt, CLIP threshold for filtering, and \% yield for each of the distribution shifts in our benchmark.}
    \label{tab:thresholds}
\end{table}

%% file: sections/appendix/results.tex
\begin{figure*}[!htb]
    \centering 
    \includegraphics[width=1\textwidth]{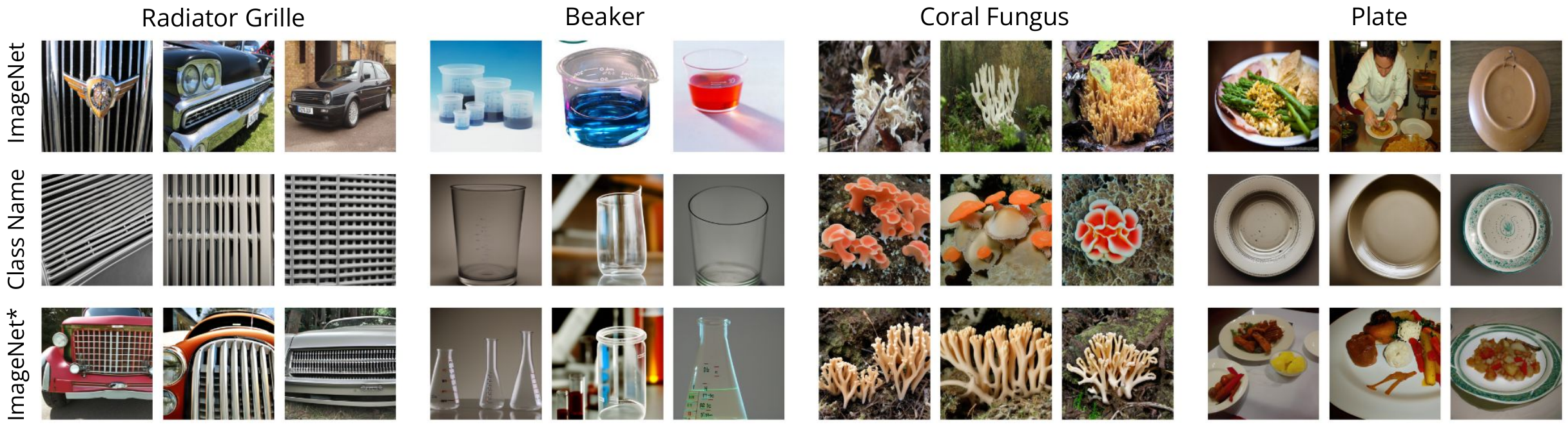}
    \caption{Additional examples of mismatch between prompting of Stable Diffusion using the class name and the ImageNet dataset. We visualize real images from ImageNet (\textbf{top}), images generated using the class name in prompts (\textbf{middle}) and ImageNet$*$ 
    (\textbf{bottom}).}
    \label{fig:extra_mismatch}
\end{figure*}
\begin{figure*}[!htb]
    \centering 
    \includegraphics[width=1\textwidth]{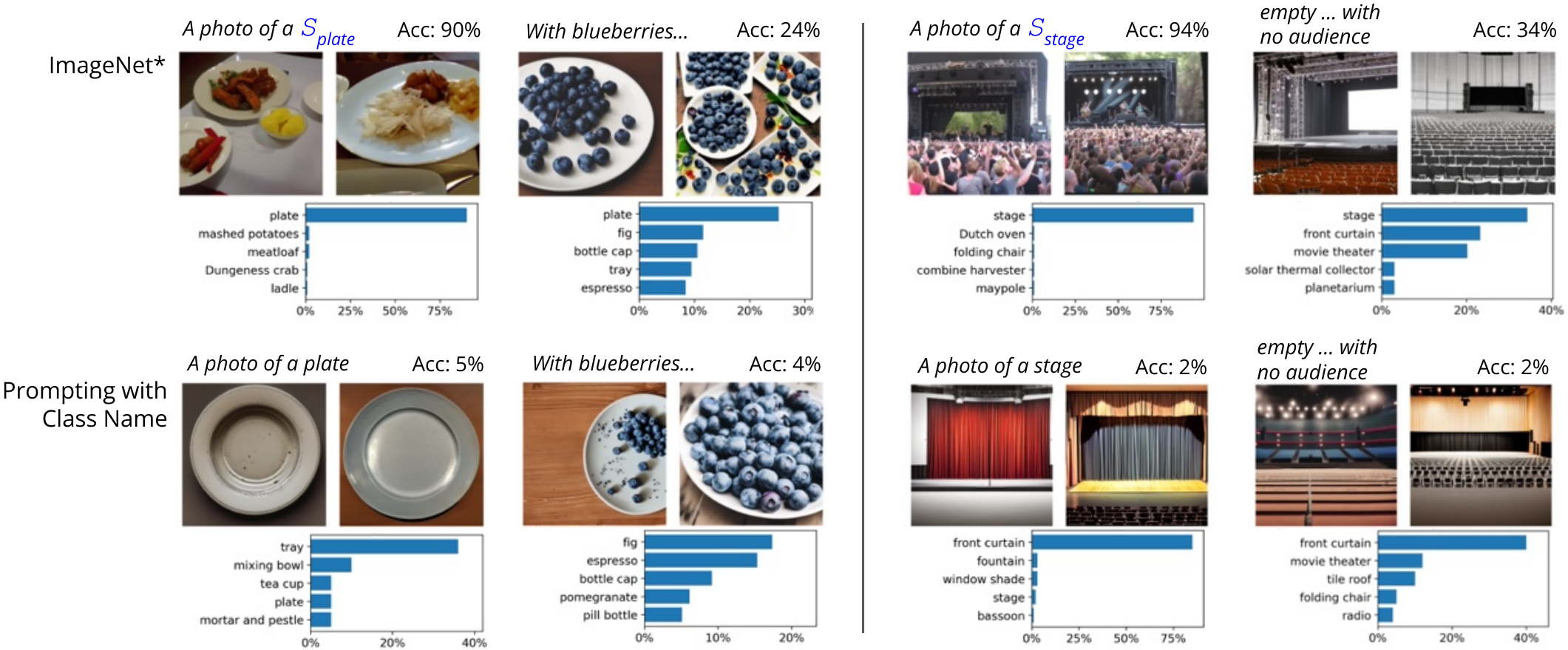}
    \caption{Additional images generated by our dataset interface ImageNet$*$ (\textbf{top}) and the corresponding images generated by prompting Stable Diffusion with the class name (\textbf{bottom}), as well as the top predicted classes of an ImageNet-trained ResNet50. Using ImageNet$*$, we find that ``plate with blueberries'' and ``empty stage with no audience'' both lead to a large degradation in the classifier's accuracy compared to base generated images. On the other hand, the images generated by Stable Diffusion when prompted with the class name all lead to low model performance regardless of the specified shift.}
    \label{fig:extra_debugging}
\end{figure*}
\begin{figure*}[!htb]
    \centering 
    \includegraphics[width=1\textwidth]{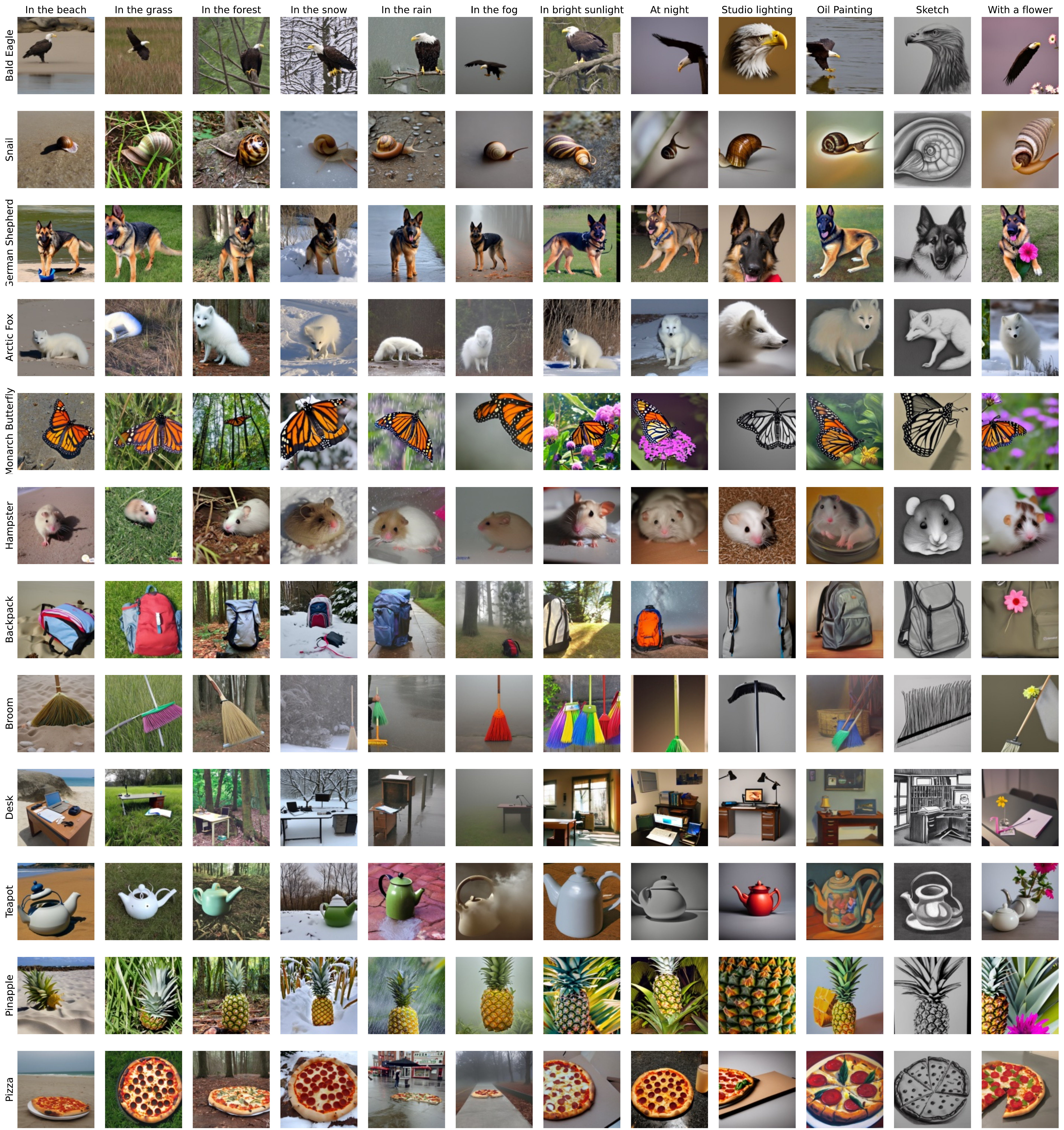}
    \caption{Additional examples of images generated with ImageNet$*$ for a variety of distribution shifts. These images are a subset of the benchmark described in Section \ref{sec:benchmark}.}
    \label{fig:extra_shifts}
\end{figure*}

Here we visualize additional experiments and examples extending upon the figures in the main paper. In Figure \ref{fig:extra_mismatch}, we show visualize further examples of classes for which there is a visual mismatch between images generated by Stable Diffusion using the natural language prompts and images in ImageNet, as in Figure \ref{fig:extra_mismatch}. In Figure \ref{fig:extra_debugging}, we show additional examples of using our dataset interface for model debugging, and as in Figure \ref{fig:debugging} we compare our counterfactual examples to those generated by prompting Stable Diffusion with natural language. In Figure \ref{fig:extra_shifts}, we extend upon the visualizations in \ref{fig:benchmark} and display additional samples of images from our distribution shift benchmark.

%% file: sections/appendix/clip.tex
Recall from Section \ref{sec:threshold} that our counterfactual examples, at the very least, must possess two properties: (1) they must contain the original object and (2) they must realistically manifest the desired shift. Therefore,  we can evaluate the quality of the generated ImageNet$*$ images with respect to both of these aspects using the metrics described in Section~\ref{sec:threshold}.

\begin{figure}
    \begin{subfigure}[b]{0.56\textwidth}
        \centering
        \includegraphics[width=1\linewidth]{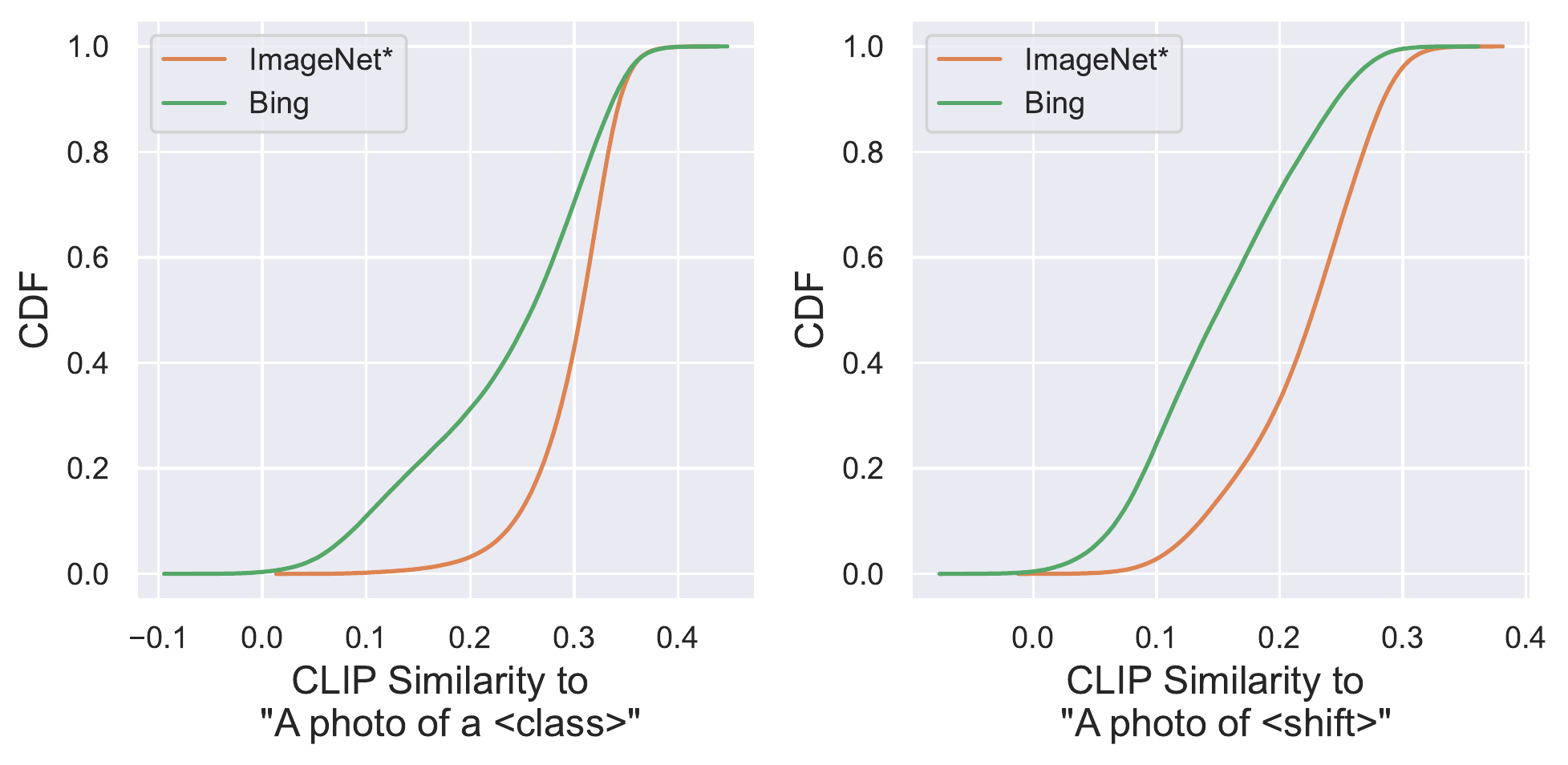}
        \subcaption{CDF of CLIP similarity to captions $c_{class}$ (\textbf{left}) and $c_{shift}$ (\textbf{right})}
        \vspace{0em}
        \label{fig:yield_cdf}
    \end{subfigure}\hfill
    \begin{subfigure}[b]{0.42\textwidth}
        \centering
        \includegraphics[width=\linewidth]{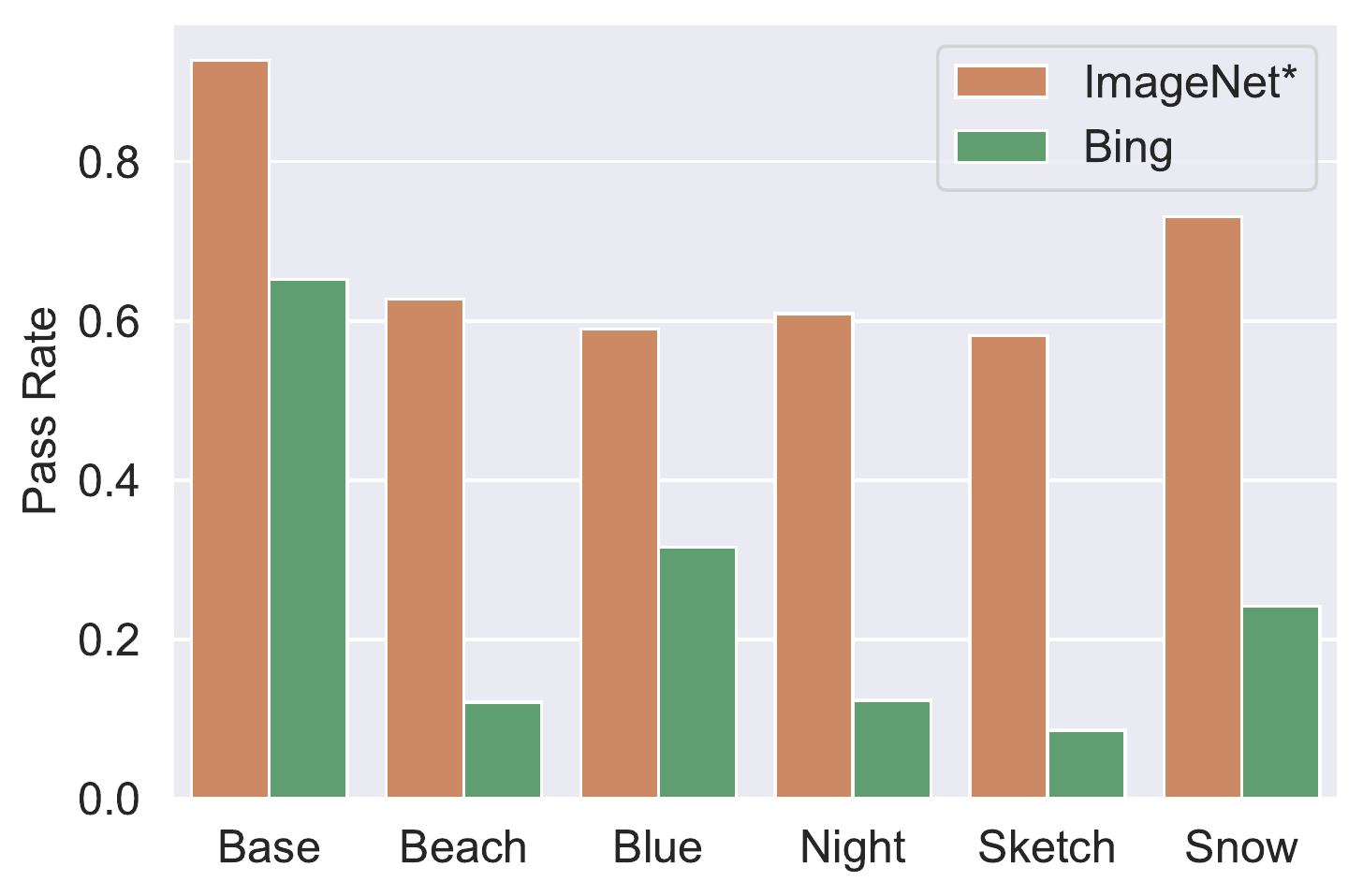}
        \subcaption{Fraction of images passing the filters.}
        \vspace{0em}
        \label{fig:barplot_yield}
    \end{subfigure}
    \caption{
        CLIP-based evaluation of the counterfactual examples produced by ImageNet$*$ for the shifts mentioned in Section~\ref{sec:distribution}. As a baseline, we also evaluate images scraped using Bing or \texttt{clip-retrieval}.
        (\textbf{a}) CDFs of the CLIP similarity between the images and the captions $c_{class}=$\textit{"A photo of a \textless class\textgreater"} and $c_{shift}=$\textit{"A photo of a \textless shift\textgreater"}. The ImageNet$*$ images turn out to have higher similarity to both captions than the scraped images.
        (\textbf{b}) The fraction of images that pass the filters described in Section~\ref{sec:method}. The ImageNet$*$ images have a consistently higher pass rate than those from Bing.
        }
\end{figure}

Specifically, we use ImageNet$*$ to synthesize images of distribution shifts from five different categories --- ``at night'', ``blue'', ``in the beach'',``in the snow'', and ``sketch'' --- as well as ``base'' images which do not correspond to a specific shift (generated with a prompt ``a photo of a S*'').
In Figure~\ref{fig:yield_cdf}, we then plot the similarity between the CLIP embedding of each  generated image and the text embedding of the captions \textit{``a photo of a \textless class\textgreater''} ($c_{class}$) and \textit{``a photo \textless shift\textgreater''} ($c_{shift}$). As a baseline, we consider downloading images returned by the Bing search engine to a natural language prompt with the ImageNet class name or retrieving them with \texttt{clip-retrieval}, an open source tool for scraping LAION. (See Appendix~\ref{app:bing} for experimental details.)

We observe that the ImageNet$*$ generated images have higher CLIP similarity to both captions than the images scraped using text-to-image retrieval methods (Figure \ref{fig:yield_cdf}). This suggests that the ImageNet$*$ images more consistently contain the original object and manifest the desired shift than their Bing counterparts. As a result, the majority of our images pass through the automatic filters described in Section~\ref{sec:threshold} (Figure~\ref{fig:barplot_yield}). 

%% file: sections/appendix/datasets.tex
In this section, we give qualitative results for applying our framework on two additional datasets, FGVC-Aircraft \cite{maji2013fine} and Oxford-IIIT Pet \cite{parkhi2012cats}. For each of these two datasets, we learn a token for each class, as for ImageNet. Due to the fine-grained nature of the datasets and the possible confusion with specific labels (such as \textit{E-170}, a type of aircraft), we initialize the Textual Inversion process for each class with the same broader description (\textit{``pet''} and \textit{``airplane''}). 

\begin{figure*}[h]
    \centering 
    \includegraphics[width=0.9\textwidth]{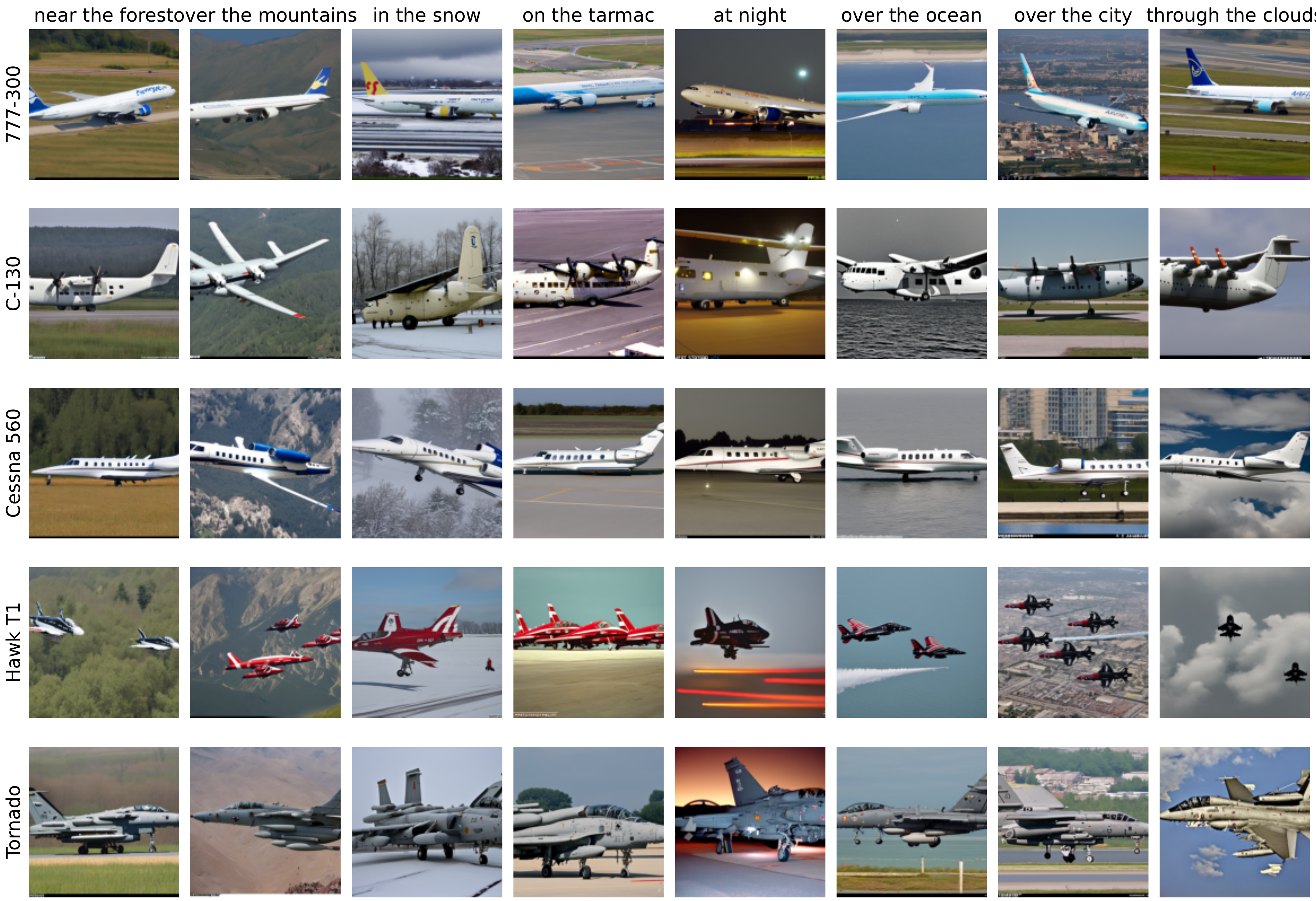}
    \caption{Examples of images generated with our learned tokens for the FGVC-Aircraft dataset in a variety of distribution shifts. These examples are not filtered with the CLIP similarity metrics.}
    \label{fig:aircraft_samples}
\end{figure*}

\begin{figure*}[!htb]
    \centering 
    \includegraphics[width=0.9\textwidth]{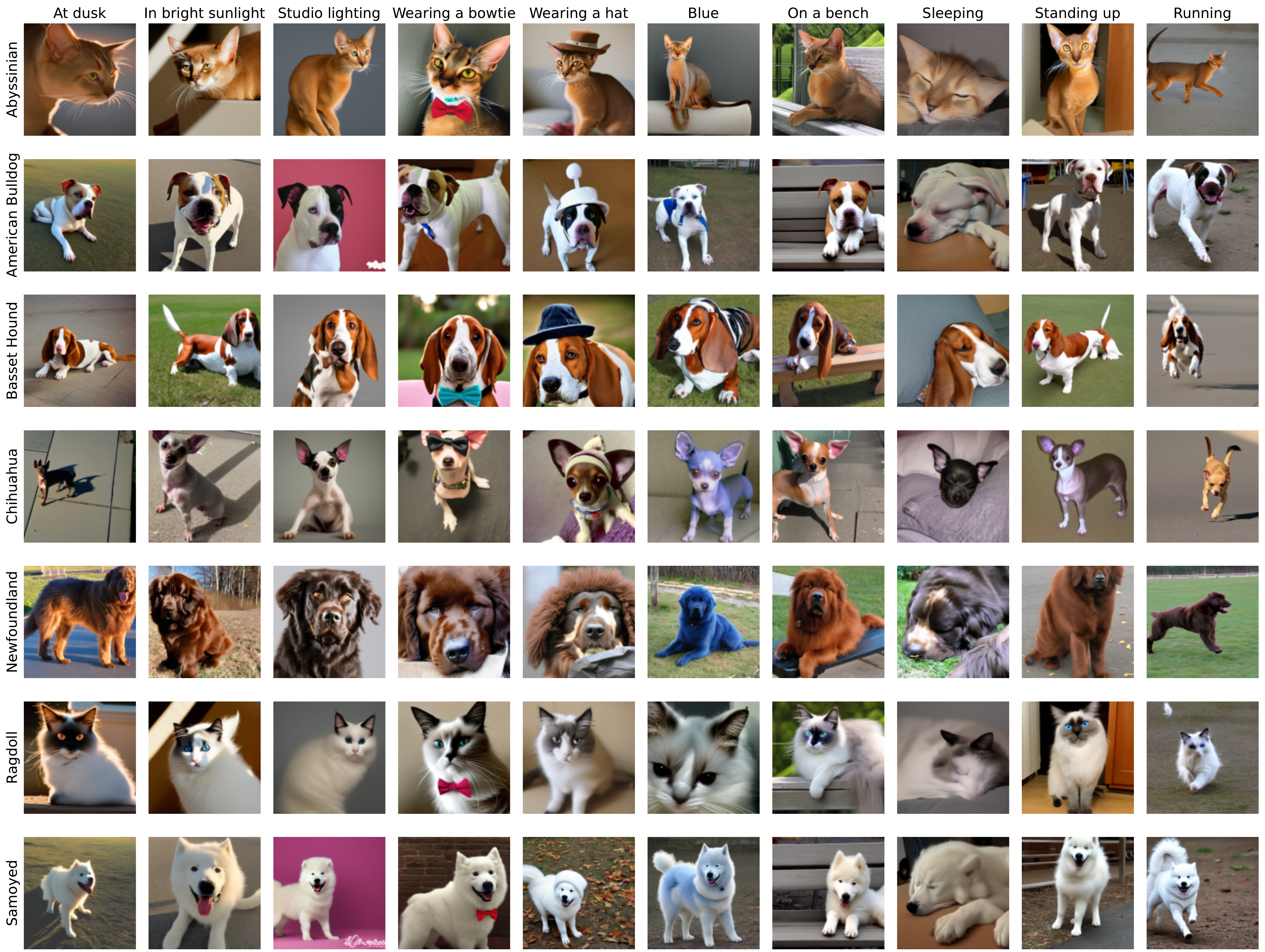}
    \caption{Examples of images generated with our learned tokens for the  Oxford-IIIT Pet dataset in a variety of distribution shifts. These examples are not filtered with the CLIP similarity metrics.}
    \label{fig:pet_samples}
\end{figure*}

In Figure \ref{fig:aircraft_samples} and \ref{fig:pet_samples}, we visualize images generated using the learned tokens in a range of distribution shifts. Due to the more specific nature of the datasets, we are able to exhibit shifts that would not apply meaningfully across all of classes of a broader dataset such as ImageNet (e.g. \textit{``through the clouds''} for FGVC-Aircraft, \textit{``sleeping''} for Oxford-IIIT Pet).

